%% file: main.tex
\documentclass[10pt,twocolumn,letterpaper]{article}

\pdfoutput=1

\usepackage{3dv}
\usepackage{times}
\usepackage{amsmath}
\usepackage{amssymb}
\usepackage{subcaption}
\usepackage{comment}
\usepackage{lipsum}
\usepackage[pdftex]{graphicx}

\usepackage[toc,page]{appendix}

\newcommand\blfootnote[1]{%
	\begingroup
	\renewcommand\thefootnote{}\footnote{#1}%
	\addtocounter{footnote}{-1}%
	\endgroup
}

\input{macros}

\threedvfinalcopy 

\def\threedvPaperID{****} 

\ifthreedvfinal\fi
\begin{document}

\title{\OURS: Learning from RGB-D Data in Indoor Environments}

\author{
	Angel Chang$^1$$^*$	\qquad
	Angela Dai$^2$$^*$ \qquad
	Thomas Funkhouser$^1$$^*$\qquad
	Maciej Halber$^1$$^*$ \\	
	\hfill
	Matthias Nie{\ss}ner$^3$$^*$ \qquad
	Manolis Savva$^1$$^*$ \qquad
	Shuran Song$^1$$^*$ \qquad
	Andy Zeng$^1$$^*$ \qquad
	Yinda Zhang$^1$$^*$
	\hfill
	\vspace{0.1cm}
	\\
	\hfill
	$^1$Princeton University \qquad
	$^2$Stanford University \qquad
	$^3$Technical University of Munich
	\hfill
	\vspace{-0.2cm}
}

\maketitle

\begin{abstract}
\blfootnote{$^*$authors are in alphabetical order}
Access to large, diverse RGB-D datasets is critical for training RGB-D scene understanding algorithms.
However, existing datasets still cover only a limited number of views or a restricted scale of spaces.
In this paper, we introduce \OURS, a large-scale RGB-D dataset containing \PANOCOUNT panoramic views from \FRAMECOUNT RGB-D images of \SCENECOUNT building-scale scenes.  Annotations are provided with surface reconstructions, camera poses, and 2D and 3D semantic segmentations.
The precise global alignment and comprehensive, diverse panoramic set of views over entire buildings enable a variety of supervised and self-supervised computer vision tasks, including keypoint matching, view overlap prediction, normal prediction from color, semantic segmentation, and region classification.
\end{abstract}

\input{1body.tex}

{\small
\bibliographystyle{ieee}
\bibliography{main}
}


\appendix
\newpage
\phantomsection
\addcontentsline{toc}{chapter}{Appendices}
\begin{huge}
\textbf{Appendix}
\end{huge}
\setcounter{section}{0}
\renewcommand{\thesection}{\Alph{section}}
\renewcommand{\theHsection}{appendixsection.\Alph{section}}

\input{0supp_body.tex}


\end{document}

%% file: macros.tex
\usepackage[usenames, dvipsnames]{color}
\newcommand{\IGNORE}[1]{{}}
\newcommand{\MATTHIAS}[1]{\textcolor{red}{{\textbf{[Matthias: #1]}}}}

\newcommand{\OURS}{Matterport3D\xspace}
\newcommand{\PANOCOUNT}{10,800\xspace}
\newcommand{\FRAMECOUNT}{194,400\xspace}
\newcommand{\SCENECOUNT}{90\xspace}

\newcommand{\NUMTASKS}{5\xspace}

\definecolor{mygreen}{RGB}{111, 51, 129}

\def \path{\bp C}



%% file: 1body.tex
\section{Introduction}

Scene understanding for RGB-D images of indoor home environments is a fundamental task for many applications of computer vision, including personal robotics, augmented reality, scene modeling, and perception assistance.

Although there has been impressive research progress on this topic, a significant limitation is the availability suitable RGB-D datasets from which models can be trained.  As with other computer vision tasks, the performance of data-driven models exceeds that of hand-tuned models and depends directly on the quantity and quality of training datasets.
Unfortunately, current RGB-D datasets have small numbers of images
\cite{silberman2012indoor}, limited scene coverage \cite{handa2015scenenet}, limited
viewpoints \cite{song2015sun}, and/or motion blurred imagery.  Most are restricted to single rooms \cite{hua2016scenenn,dai2017scannet}, synthetic imagery \cite{handa2015scenenet,song2016semantic}, and/or a relatively small number of office environments \cite{armeni2017joint}. 
No previous dataset provides high-quality RGB-D images for a diverse set of views in interior home environments.

\begin{figure}[t]
	\centering
	\includegraphics[width=\linewidth]{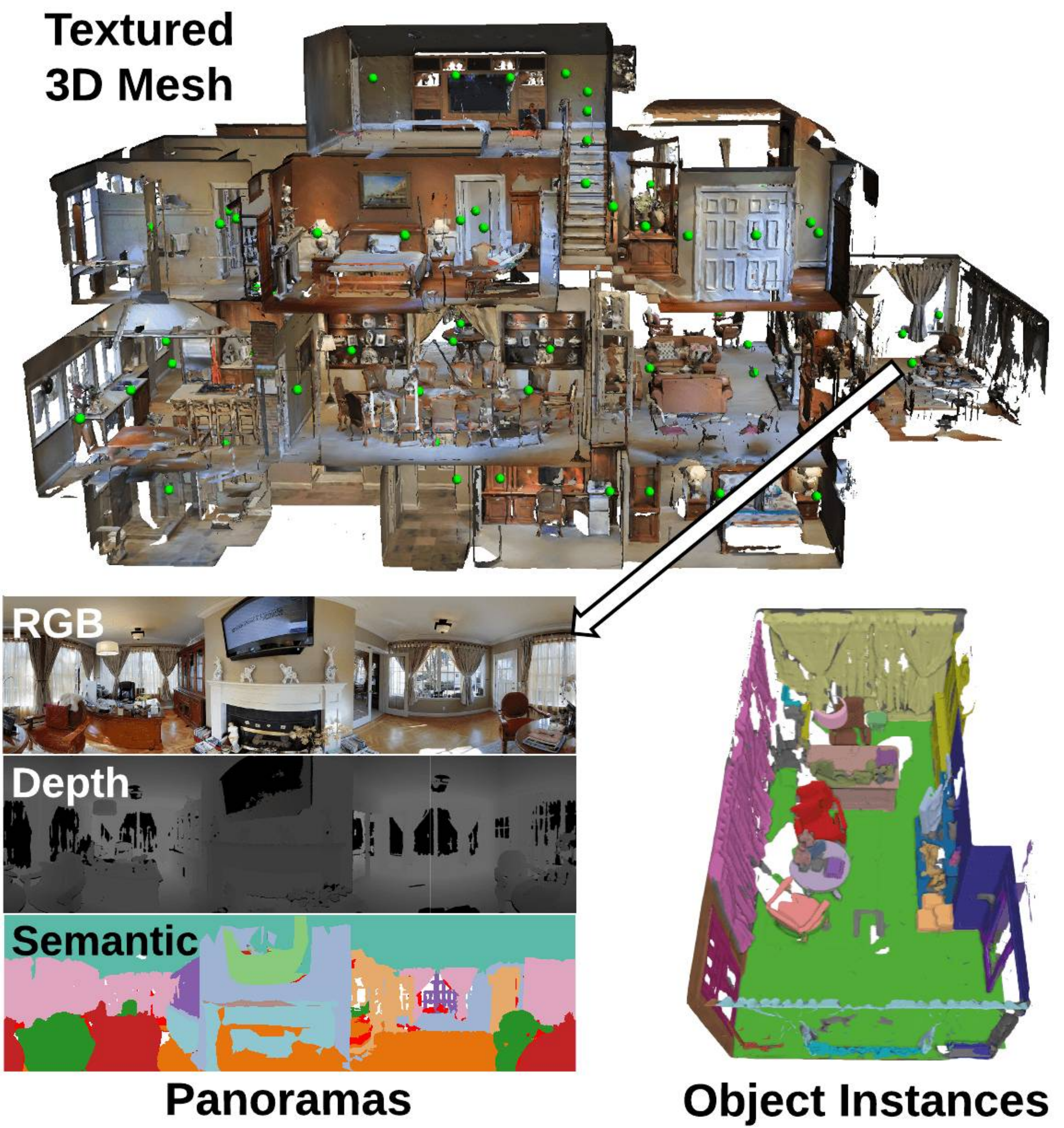}
	\caption{The \OURS dataset provides visual data covering \SCENECOUNT buildings, including HDR color images, depth images, panoramic skyboxes, textured meshes, region layouts and categories, and object semantic segmentations.\vspace{-0.2cm}}
	\label{fig:teaser}
\end{figure}

This paper introduces the \OURS dataset and investigates new research opportunities it provides for learning about indoor home environments.
The dataset comprises a set of \FRAMECOUNT RGB-D images captured in \PANOCOUNT panorama with a Matterport camera\footnote{\url{https://matterport.com/}} in  home environments. 
Unlike previous datasets, it includes both depth and color $360^\circ$ panoramas for each viewpoint, samples human-height viewpoints uniformly throughout the entire environment, provides camera poses that are globally consistent and aligned with a textured surface reconstruction, includes instance-level semantic segmentations into region and object categories, and provides data collected from living spaces in private homes.

Though the curation of the dataset is interesting in its own, the most compelling part of the project is the computer vision tasks enabled by it.   
In this paper, we investigate \NUMTASKS tasks, each leveraging different properties of the dataset.
The precise global alignment over building scale allows training for state-of-the-art keypoint descriptors that can robustly match keypoints from drastically varying camera views. 
The panoramic and comprehensive viewpoint sampling provides a large number of loop closure instances, allowing learning of loop closure detection through predicting view overlap.
The surface normals estimated from high-quality depths in diverse scenes allows training models for normal estimation from color images that outperform previous ones.
The globally consistent registration of images to a surface mesh facilitates semantic annotation, enabling efficient 3D interfaces for object and region category annotation from which labels projected into images can train deep networks for semantic segmentation.
For each of these tasks, we provide baseline results using variants of existing state-of-the-art algorithms demonstrating the benefits of the \OURS data; we hope that \OURS will inspire future work on many scene understanding tasks\footnote{All data and code is publicly available:\\ \url{https://github.com/niessner/Matterport}}.

\section{Background and Related Work}

Collecting and analyzing RGB-D imagery to train algorithms for scene understanding is an active area of research with great interest in computer vision, graphics, and robotics \cite{firman2016rgbd}.
Existing work on curation of RGB-D datasets has focused mostly on scans of individual objects \cite{choi2016large}, stand-alone rooms \cite{hua2016scenenn,savva2016pigraphs}, views of a room \cite{silberman2011indoor,song2015sun}, spaces from academic buildings and small apartments \cite{dai2017scannet}, and small collections of rooms or public spaces \cite{xiao2013sun3d,armeni2017joint,knapitsch2017tanks}. 
Some of these datasets provide 3D surface reconstructions and object-level semantic annotations \cite{hua2016scenenn,armeni20163d,savva2016pigraphs,dai2017scannet}.
However, none have the scale, coverage, alignment accuracy, or HDR imagery of the dataset presented in this paper.  

These previous RGB-D datasets have been used to train models for several standard scene understanding tasks, including semantic segmentation \cite{ren2012rgb,silberman2012indoor,gupta2013perceptual,valentin2015semanticpaint,dai2017scannet}, 
3D object detection \cite{shrivastava2013building,lin2013holistic,gupta2014learning,song2014sliding,song2015deep}, 
normal estimation \cite{wang2015designing,li2015depth,eigen2015predicting,bansal2016marr},
camera relocalization \cite{shotton2013scene,valentin2016learning},
and others \cite{zhang2013estimating,fouhey2013data,fouhey2014unfolding}.
We add to this body of work by investigating tasks enabled by our dataset, including 
learning an image patch descriptor, predicting image overlaps, estimating normals, 
semantic voxel labeling, and classifying images by region category.

The work most closely related to ours is by Armeni et al. \cite{armeni2017joint}.
They also utilize a 3D dataset collected with Matterport cameras for scene understanding.  
However, there are several important differences.
First, their data is collected in only 3 distinct office buildings, whereas we have data from \SCENECOUNT distinct buildings with a variety of scene types including homes (mostly), offices, and churches. 
Second, their dataset contains only RGB images and a coarse surface mesh from which they generate a point cloud -- we additionally provide the raw depth and HDR images collected by Matterport. 
Third, their semantic annotations cover only 13 object categories, half of which are structural building elements -- we collect an open set of category labels which we reduce to 40 categories with good coverage of both building elements and objects.  
Finally, their algorithms focus only on tasks related to semantic parsing of buildings into spaces and elements, while we consider a wide range of tasks enabled by both supervised and self-supervised learning.

\section{The \OURS Dataset}

This paper introduces a new RGB-D dataset of building-scale scenes, and describes a set of scene understanding tasks that can be trained and tested from it.  
We describe the data in this section, along with a discussion of how it differs from prior work.  

\subsection{Data Acquisition Process}
\label{sec:acquisition}

The Matterport data acquisition process uses a tripod-mounted camera rig with three color and three depth cameras pointing slightly up, horizontal, and slightly down.  
For each panorama, it rotates around the direction of gravity to 6 distinct orientations, stopping at each to acquire an HDR photo from each of the 3 RGB cameras.  
The 3 depth cameras acquire data continuously as the rig rotates, which is
integrated to synthesize a 1280x1024 depth image aligned with each color image.
The result for each panorama is 18 RGB-D images with nearly coincident centers of projection at approximately the height of a human observer.

\begin{figure}[t]
	\includegraphics[width=\linewidth]{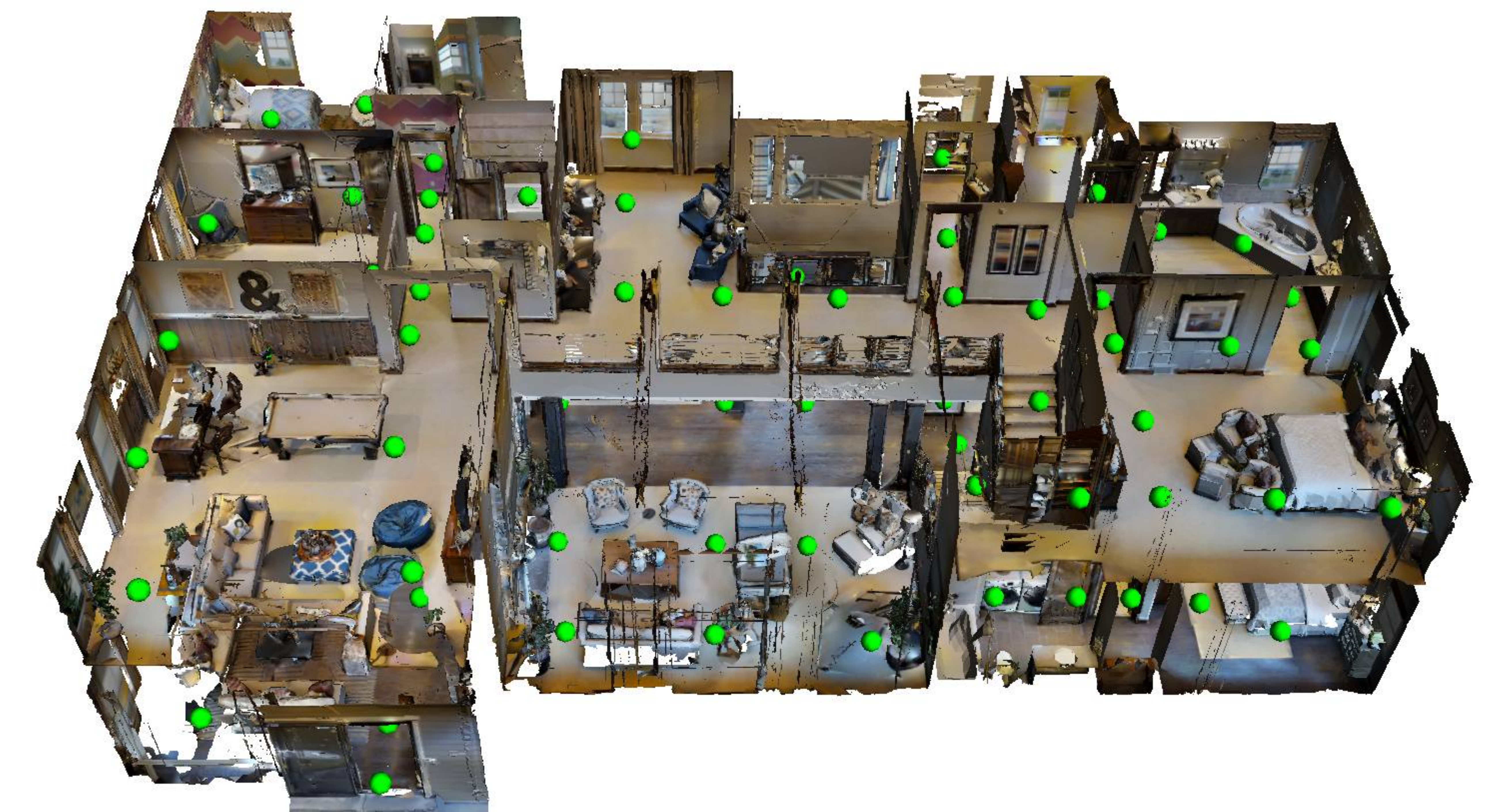}
	\caption{Panoramas are captured from viewpoints (green spheres) on average 2.25m apart.\vspace{-0.2cm}}
	\label{fig:matterport_dataset}
\end{figure}

For each environment in the dataset, an operator captures a set of panoramas uniformly spaced at approximately 2.5m throughout the entire walkable floor plan of the environment (Figure \ref{fig:matterport_dataset}). 
The user tags windows and mirrors with an iPad app and uploads the data to Matterport.  
Matterport then processes the raw data by: 
1) stitching the images within each panorama into a ``skybox'' suitable
for panoramic viewing, 
2) estimating the 6 DoF pose for each image with global bundle adjustment, 
and 3) reconstructing a single textured mesh containing all visible surfaces of the environment.

The result of this process for each scene is a set of RGB-D images at 1280x1024 (with color in HDR) with a 6 DoF camera pose estimate for each, plus a skybox for each group of 18 images in the same panorama, and a textured mesh for the entire scene.
In all, the dataset includes \SCENECOUNT buildings containing a total of \FRAMECOUNT RGB-D images, \PANOCOUNT panorama, 
and 24,727,520 textured triangles; we provide textured mesh reconstructions obtained with \cite{kazhdan2006poisson} and \cite{niessner2013hashing}.

\subsection{Semantic Annotation}
\label{sec:semantic_annotation}



We collect instance-level semantic annotations in 3D by first creating a floor plan annotation for each house, extracting room-like regions from the floor plan, and then using a crowdsourced painting interface to annotate object instances within each region.

\begin{figure}[t]
	\includegraphics[width=0.49\linewidth]{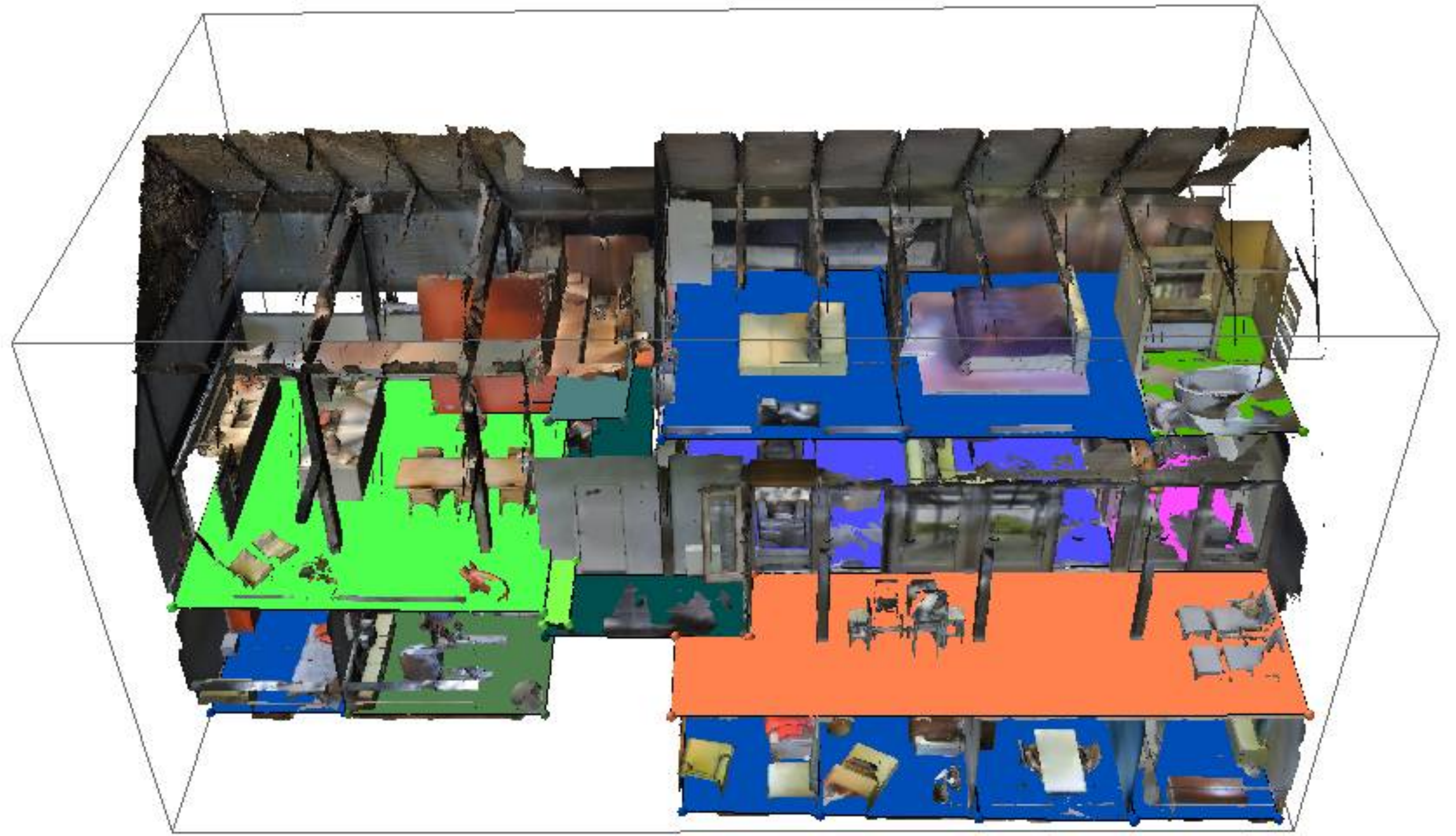}
	\includegraphics[width=0.49\linewidth]{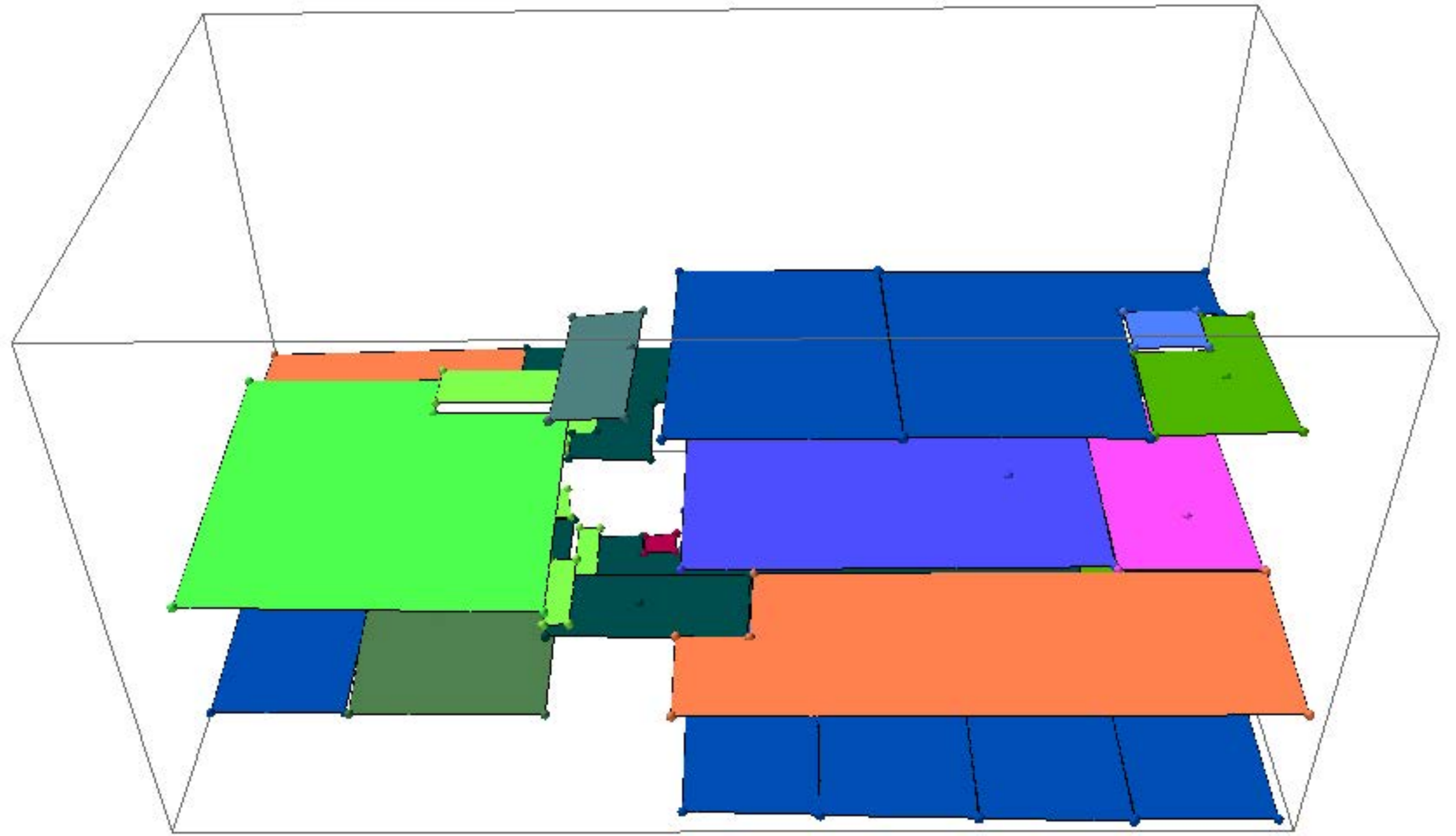}
	\caption{Annotator-specified floor plans. Floor plans are used to define regions for object-level semantic annotation.  Left: floor plan with textured mesh.  Right: floor plan alone (colored by region category).\vspace{-0.2cm}}
	\label{fig:floorplan}
\end{figure}

The first step of our semantic annotation process is to break down each building into region components by specifying the 3D spatial extent and semantic category label for each room-like region.  Annotators use a simple interactive tool in which the annotator selects a category and draws a 2D polygon on the floor for each region (see Figure \ref{fig:floorplan}).  The tool then snaps the polygon to fit planar surfaces (walls and floor) and extrudes it to fit the ceiling.

\begin{figure}[t]
	\centering
	\begin{subfigure}[t]{0.3\linewidth}
		\centering
		\includegraphics[width=\linewidth]{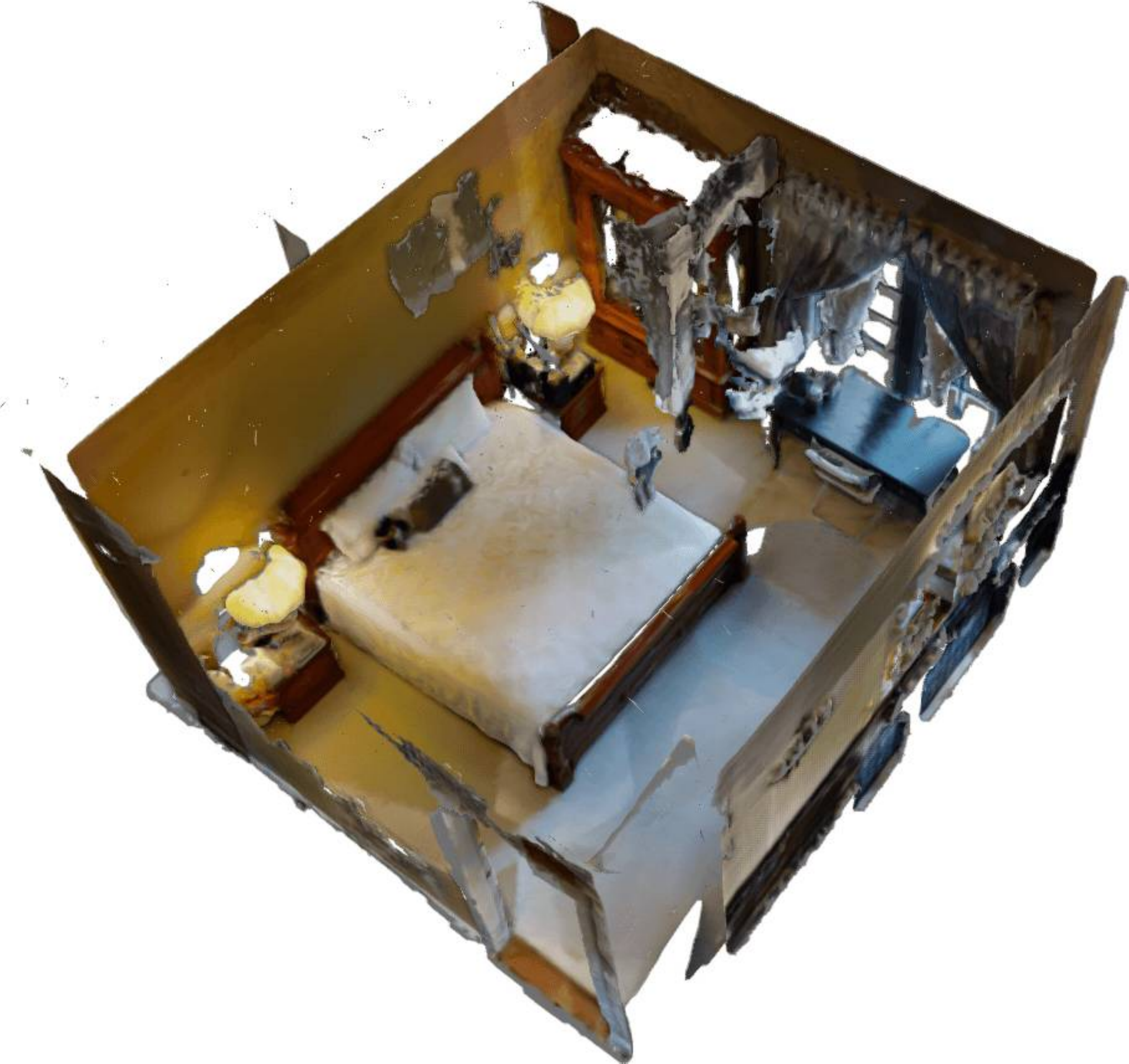}
	\end{subfigure}
	\begin{subfigure}[t]{0.3\linewidth}
		\centering
		\includegraphics[width=\linewidth]{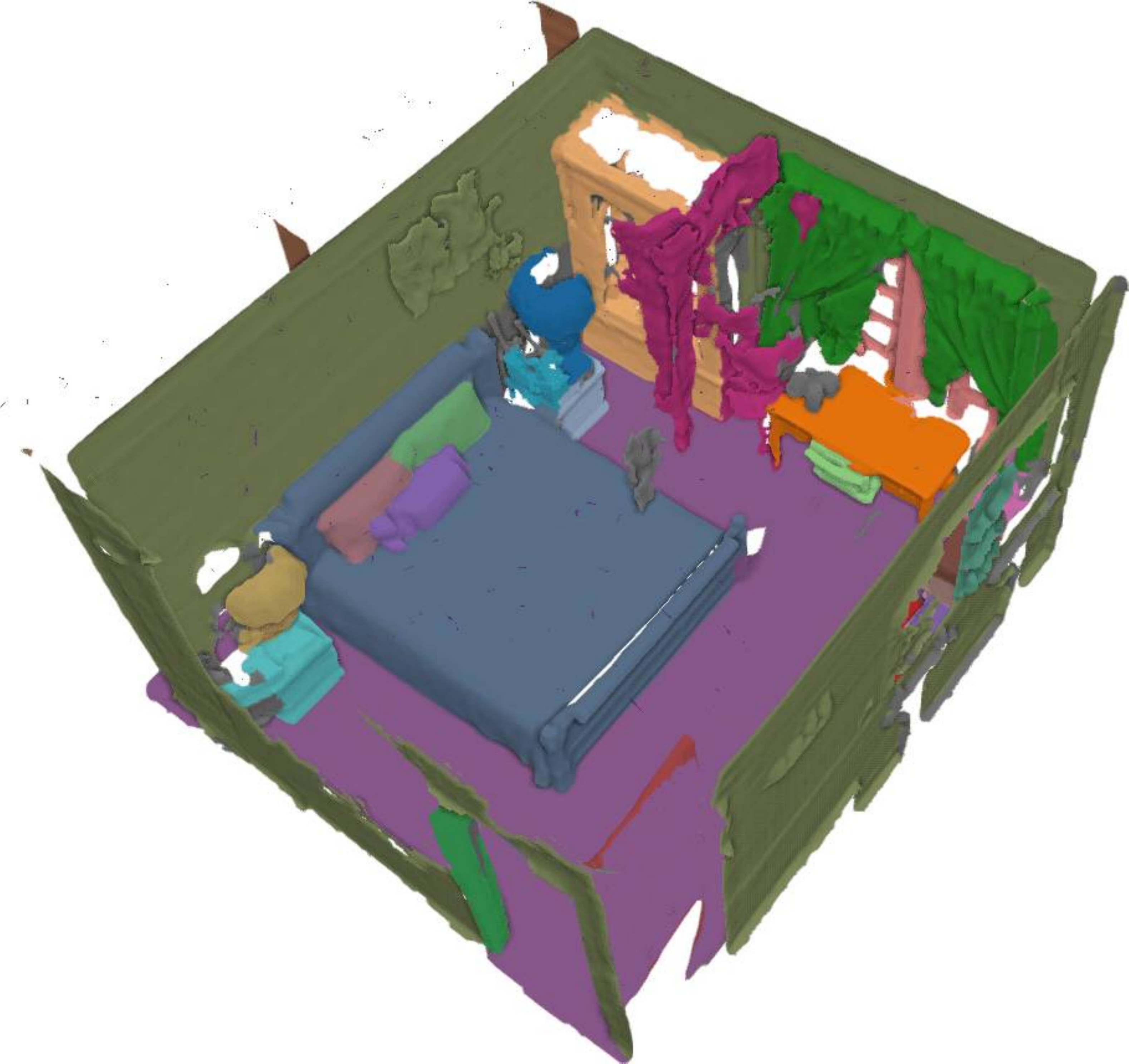}
	\end{subfigure}
	\begin{subfigure}[t]{0.3\linewidth}
		\centering
		\includegraphics[width=\linewidth]{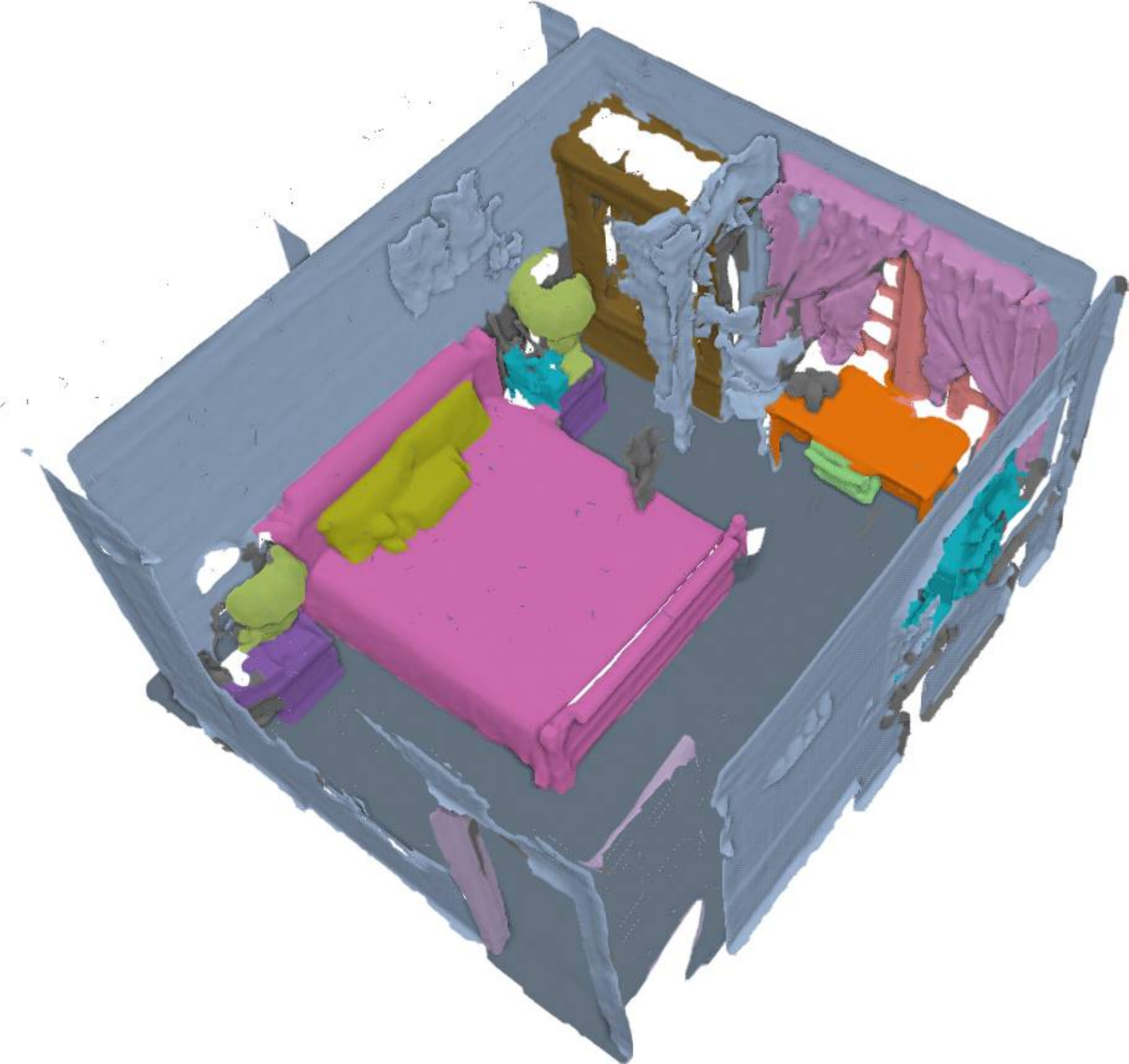}
	\end{subfigure}
	\vskip\baselineskip
	\begin{subfigure}[t]{0.27\linewidth}
		\centering
		\includegraphics[width=\linewidth]{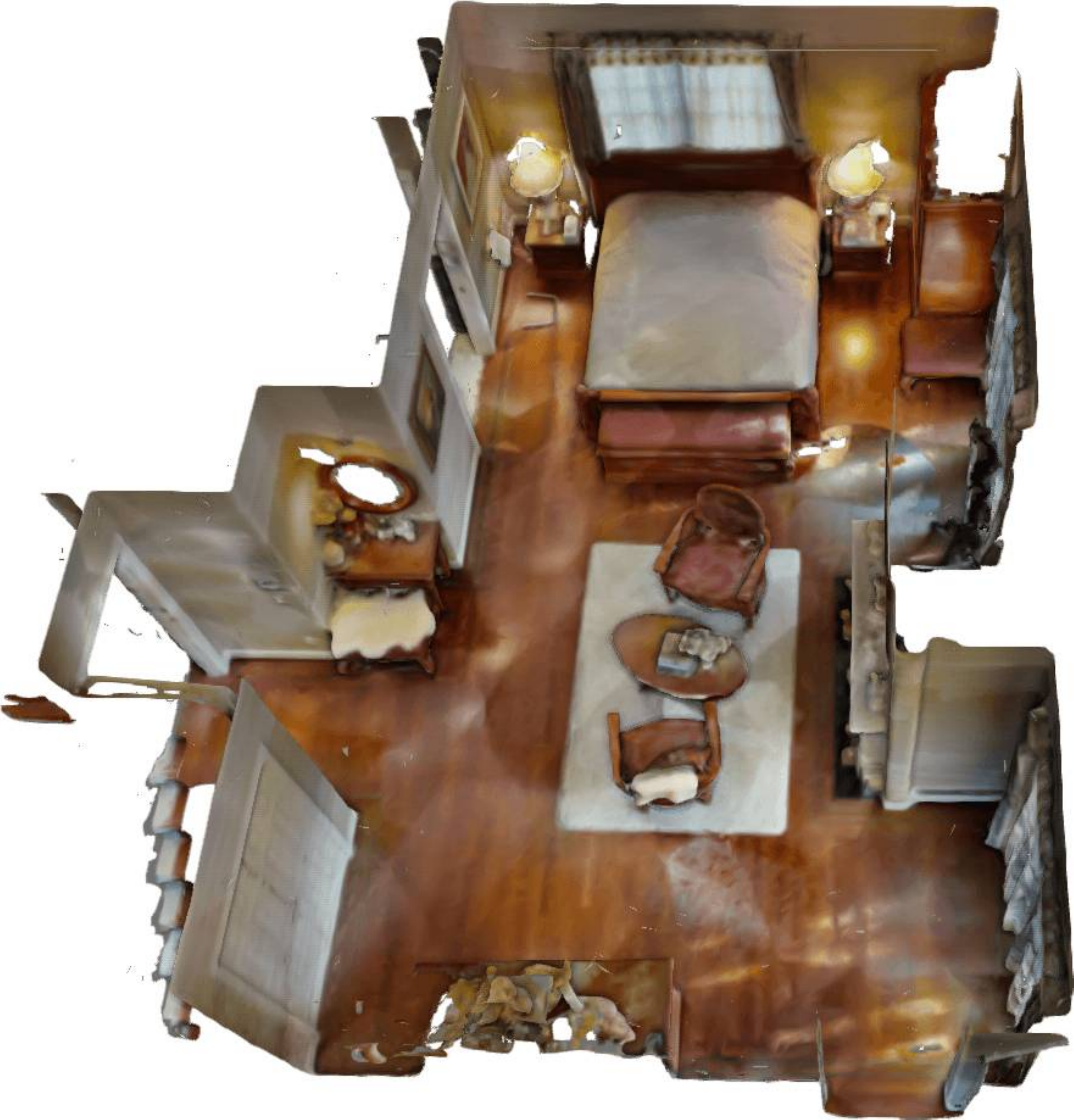}
	\end{subfigure}
	\quad
	\begin{subfigure}[t]{0.27\linewidth}
		\centering
		\includegraphics[width=\linewidth]{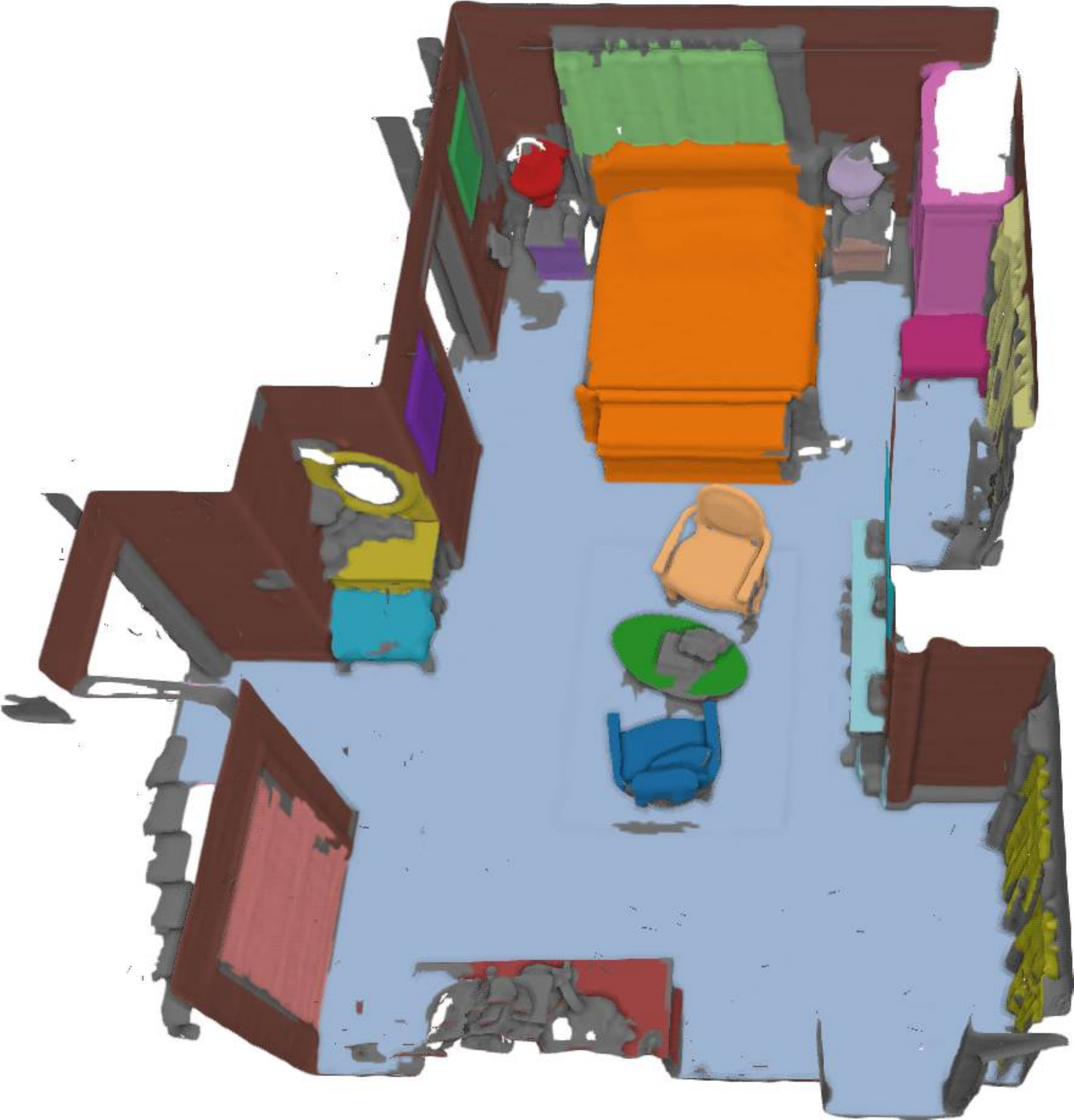}
	\end{subfigure}
	\quad
	\begin{subfigure}[t]{0.27\linewidth}
		\centering
		\includegraphics[width=\linewidth]{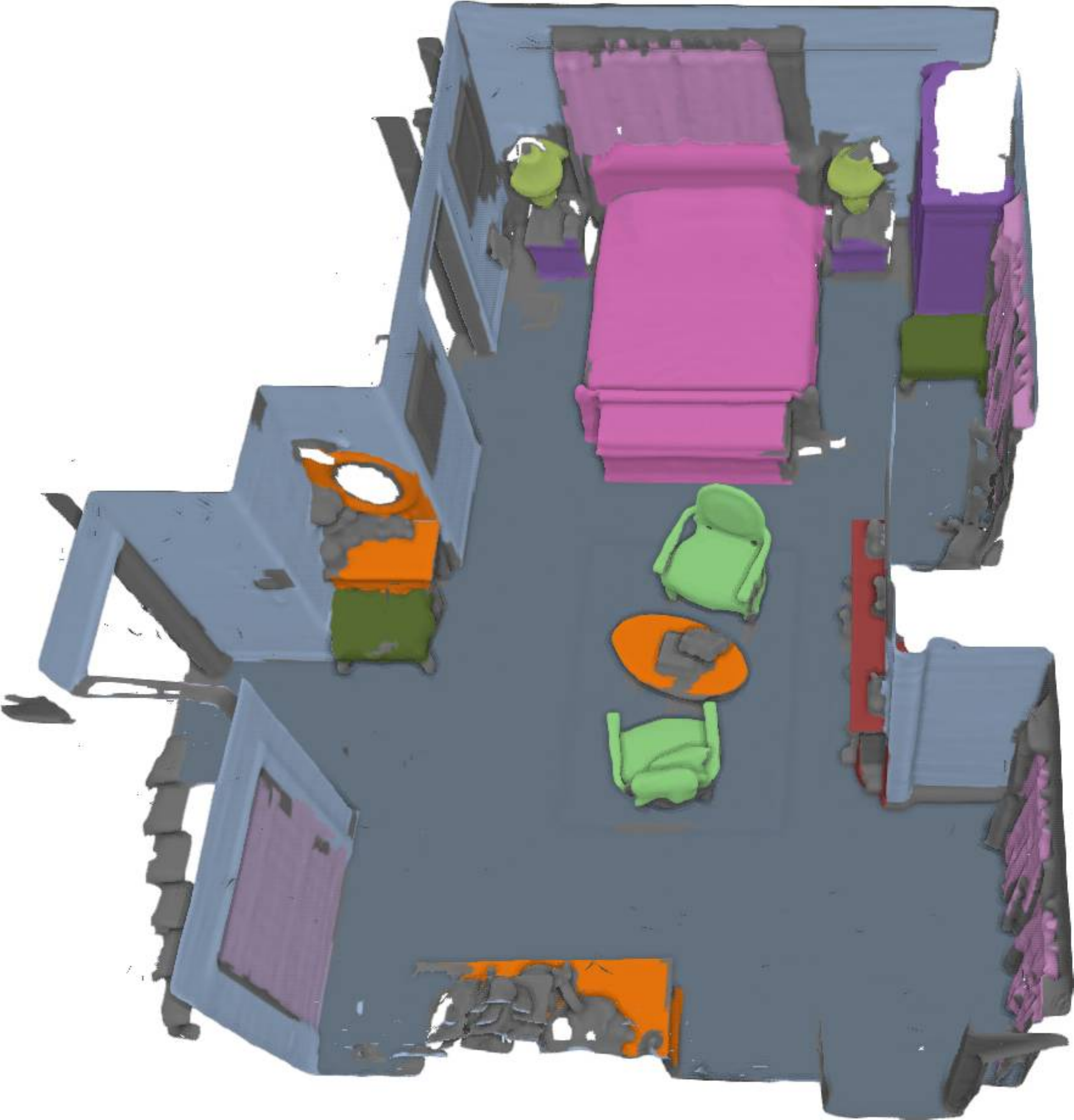}
	\end{subfigure}
	\caption{Instance-level semantic annotations. Example rooms annotated with semantic categories for all object instances. Left: 3D room mesh. Middle: object instance labels. Right: object category labels.\vspace{-0.2cm}}
	\label{fig:annotation_examples}
\end{figure}

\begin{figure*}[t]
	\includegraphics[width=\linewidth]{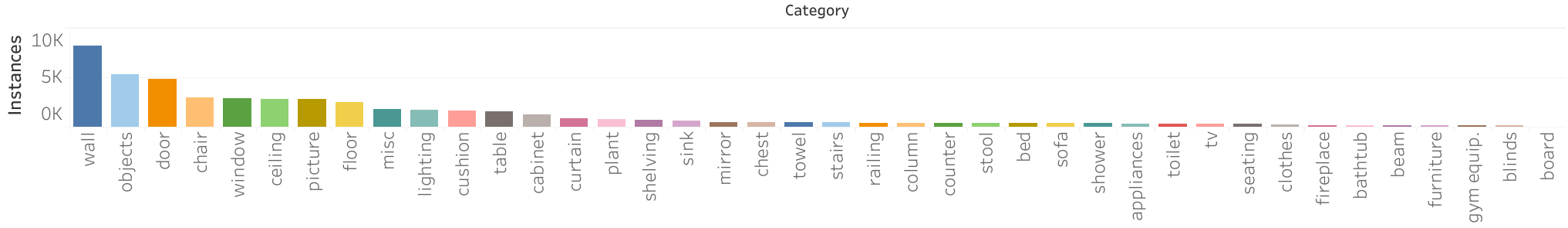}
	\vspace{-0.8cm}
	\caption{Semantic annotation statistics. Total number of semantic annotations for the top object categories.\vspace{-0.3cm}}
	\label{fig:object_category_annotations}
\end{figure*}

The second step is to label 3D surfaces on objects in each region.   
To do that, we extract a mesh for each region using screened Poisson surface reconstruction \cite{chuang2011interactive}.  
Then, we use the ScanNet crowd-sourcing interface by Dai et al.~\cite{dai2017scannet} to ``paint'' triangles to segment and name all object instances within each region.  
We first collect an initial set of labels on Amazon Mechanical Turk (AMT), which we complete, fix, and verify by ten expert annotators. We ensure high-quality label standards, as well as high annotation coverage.

The 3D segmentations contain a total of 50,811 object instance annotations.
Since AMT workers are allowed to provide freeform text labels, there were 1,659 unique
text labels, which we then post-processed to establish a canonical set of 40 object categories mapped to WordNet synsets.
Figure~\ref{fig:object_category_annotations} shows the distribution of objects by semantic category and Figure~\ref{fig:annotation_examples} shows some examples illustrated as colored meshes.

\subsection{Properties of the Dataset}
\label{sec:properties}

In comparison to previous datasets, \OURS has unique properties that open up new research opportunities:

\vspace{2mm}
\noindent{\bf RGB-D Panoramas.}  Previous panorama datasets have provided
either no depths at all \cite{xiao2012recognizing} or approximate depths synthesized
from meshes \cite{armeni2017joint}. 
\OURS contains aligned 1280x1024 color and depth images for 18 viewpoints covering approximately 3.75sr (the entire sphere except the north and south poles), along with ``skybox'' images reconstructed for outward looking views aligned with sides of a cube centered at the panorama center.  
These RGB-D panorama provide new opportunities for recognizing scene categories, estimating region layout, learning contextual
relationships, and more (see Section \ref{sec:region_classification}).

\begin{figure}[t]
	\includegraphics[width=\linewidth]{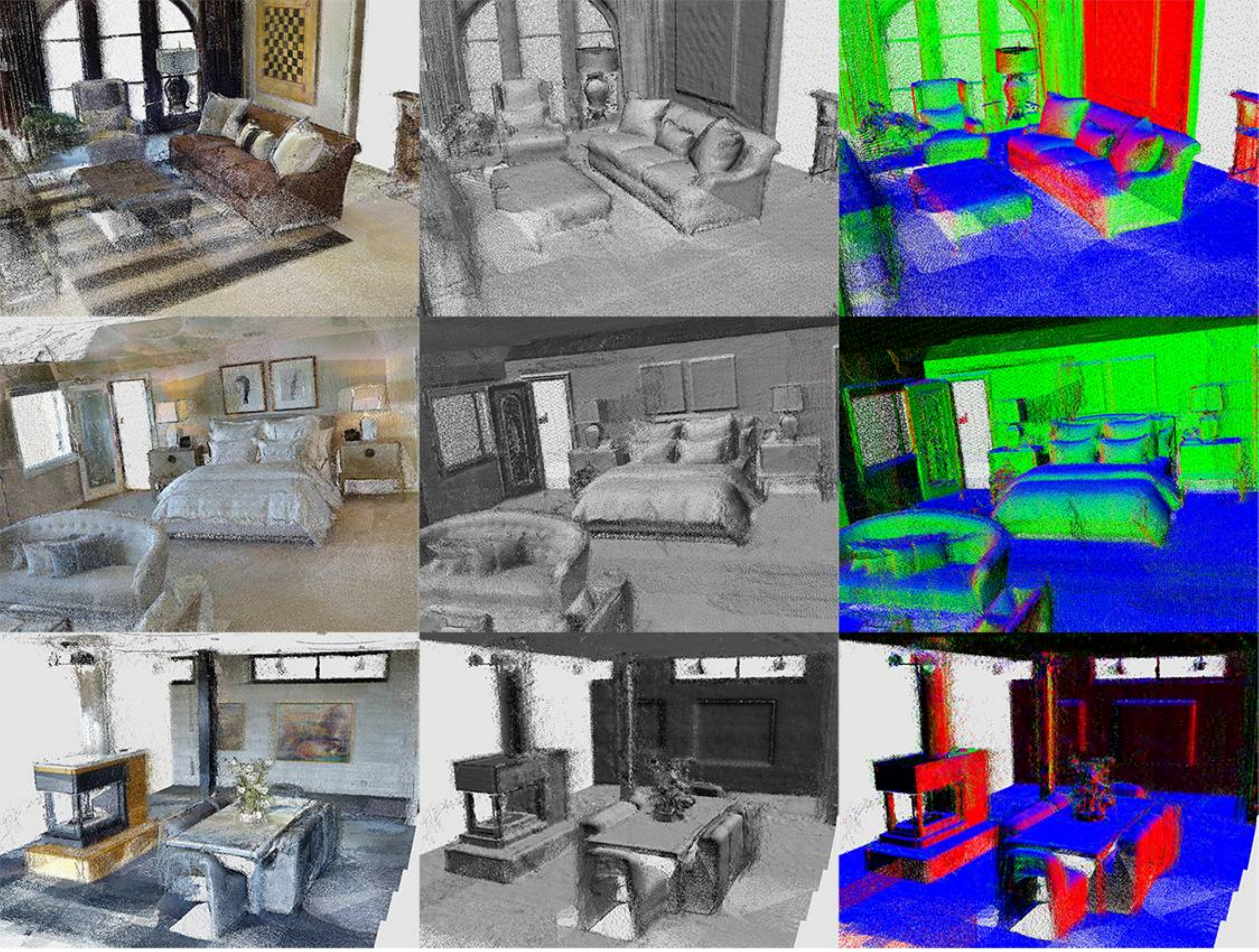}
	\caption{Visualizations of point clouds (left-to-right: color, diffuse shading, and normals). These images show pixels from all RGB-D images back-projected into world space according to the provided camera poses.   Please note the accuracy of the global alignment (no ghosting) and the relatively low noise in surface normals, even without advanced depth-fusion techniques.\vspace{-0.2cm}}
	\label{fig:room_vis}
\end{figure}

\vspace{2mm}
\noindent{\bf Precise Global Alignment.}  Previous RGB-D datasets have provided limited data about global alignment of camera poses.  
Some datasets targeted at SLAM applications \cite{dai2017bundle} provide tracked camera
poses covering parts of rooms \cite{shotton2013scene} or estimated camera poses
for individual rooms \cite{dai2017scannet}, and Armeni et al.~\cite{armeni20163d} provides globally-registered camera poses for 6 floors of 3 buildings.  
Ours provides global registered imagery covering all floors of \SCENECOUNT reconstructed buildings.  
Although we do not have ground-truth camera poses for the dataset and so cannot measure errors
objectively, we subjectively estimate that the average registration error between corresponding surface points is ~1cm or less (see Figure \ref{fig:room_vis}).  
There are some surface misalignments as large as 10cm or more, but they are rare and usually for pairs of images whose viewpoints are separated by several meters.


\vspace{2mm}
\noindent{\bf Comprehensive Viewpoint Sampling.}  Previous datasets have
contained either a small set of images captured for views around ``photograph viewpoints'' \cite{song2015sun} or a sequence of video images aimed at up-close scanning of surfaces \cite{dai2017scannet}.  
Ours contains panoramic images captured from a comprehensive, sparse sampling of viewpoint space.  
Panoramic images are spaced nearly uniformly with separations of 2.25m $\pm$ 0.57m, and thus
most plausible human viewpoints are within 1.13m of a panorama center.
This comprehensive sampling of viewpoint space provides new opportunities for learning about scenes as seen from arbitrary viewpoints that may be encountered by robots or wearable sensors as they navigate through them (see Section \ref{sec:overlap_prediction}).

\vspace{2mm}
\noindent{\bf Stationary Cameras.}  Most RGB-D image datasets have
been captured mostly with hand-held video cameras and thus suffer from
motion blur and other artifacts typical of real-time scanning; e.g., pose errors, color-to-depth misalignments, 
and often contain largely incomplete scenes with limited coverage.
Our dataset contains high dynamic range (HDR) images acquired in static scenes from stationary cameras mounted 
on a tripod, and thus has no motion blur.   This property provides new opportunities to study fine-scale 
features of imagery in scenes, for example to train very precise keypoint or boundary detectors.

\begin{figure}[t]
	\includegraphics[width=\linewidth]{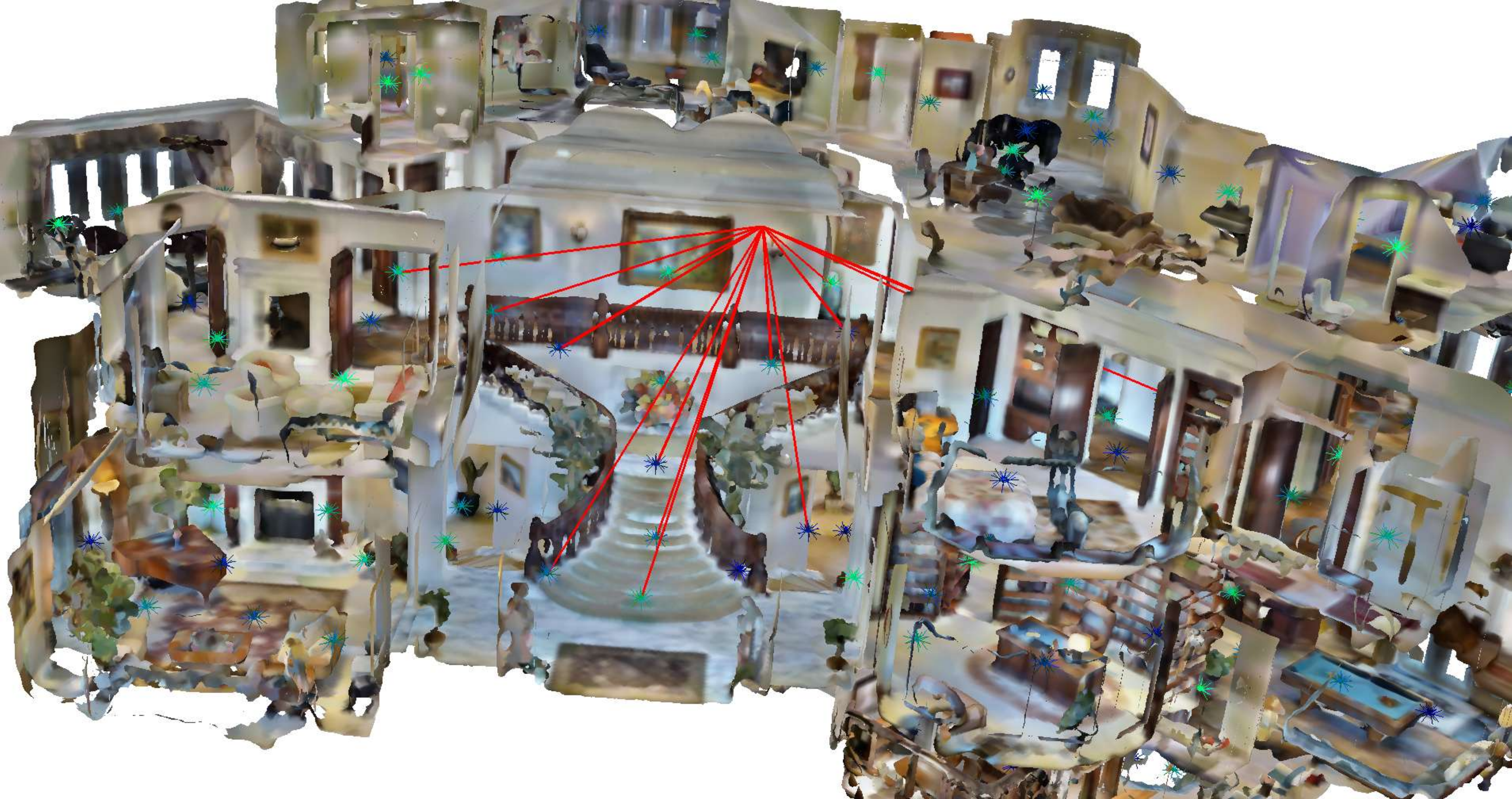}
	\caption{Visualization of the set of images visible to a selected surface point (shown as red visibility lines).  \em{(Please note that the mesh is highly decimated in this image for convenience of visualization)}\vspace{-0.5cm}}
	\label{fig:vi_overlaps}
\end{figure}

\begin{figure}[t]
	\includegraphics[width=\linewidth]{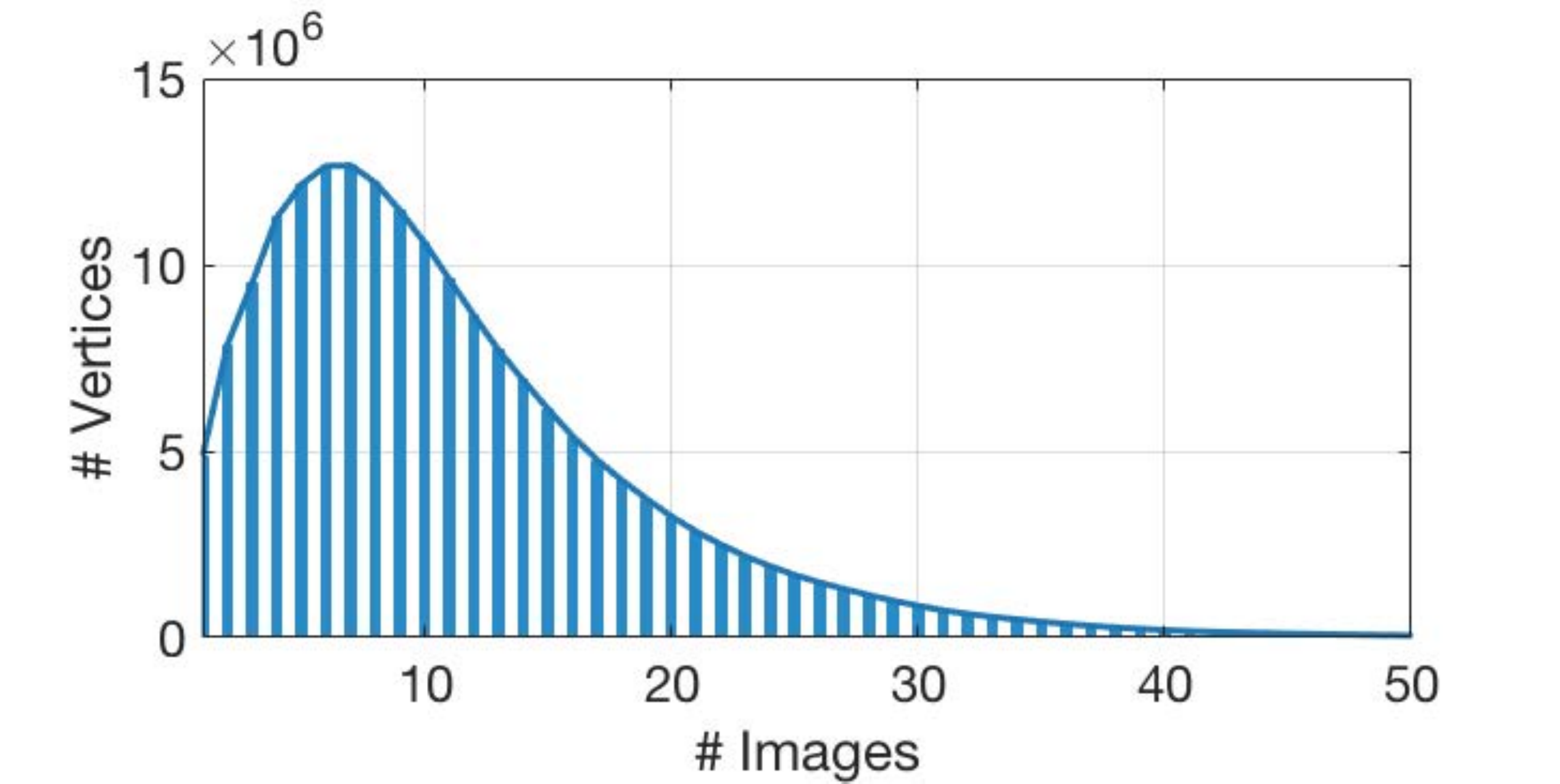}
	\caption{Histogram showing how many images observe each surface vertex.  
		The mode is 7 and the average is 11.}
	\label{fig:images_per_vertex}
\end{figure}

\vspace{2mm}
\noindent{\bf Multiple, Diverse Views of Each Surface.} 
Previous RGB-D datasets have provided a limited range of views for
each surface patch.  
Most have expressly attempted to cover each surface patch once, either to improve the efficiency of scene reconstruction or to reduce bias in scene understanding datasets.
Ours provides multiple views of surface patches from a wide variety of
angles and distances (see Figure~\ref{fig:vi_overlaps}).  
Each surface patch is observed by 11 cameras on average (see Figure~\ref{fig:images_per_vertex}).  
The overall range of depths for all pixels has mean 2.125m and standard deviation 1.4356m,
and the range of angles has mean $\text{42.584}^\circ$ and standard deviation $\text{15.546}^\circ$.  
This multiplicity and diversity of views enables opportunities for learning to predict view-dependent surface properties, such as material reflectance \cite{bell14intrinsic,rematas2016deep}, and for learning
to factor out view-dependence when learning view-independent representations, such as patch descriptors \cite{yi2016lift,zeng20163dmatch} and normals \cite{eigen2015predicting,li2015depth,bansal2016marr,wang2015designing,zhang2016physically} (see Section \ref{sec:normal_estimation}).

\vspace{2mm}
\noindent{\bf Entire Buildings.}  Previous RGB-D datasets have provided data for single rooms or small sets of adjacent
rooms \cite{xiao2013sun3d,dai2017scannet}, or single floors of a building \cite{armeni20163d}.  
Ours provides data for \SCENECOUNT entire buildings.  
On average, each scanned building has 2.61 floors, covers 2437.761$\text{m}^2$ of surface area, and has 517.34$\text{m}^2$ of floorspace.  
Providing scans of homes in their entirety enables opportunities for learning about long-range context, which is critical for holistic scene understanding and autonomous navigation.

\vspace{2mm}
\noindent{\bf Personal Living Spaces.} Previous RGB-D datasets are often limited to academic buildings \cite{armeni2017joint}. 
Ours contains imagery acquired from 
private homes (with permissions to distribute them for academic research).
Data of this type is difficult to capture and distribute due to privacy concerns, and thus it is very valuable for learning about the types of the personal living spaces targeted by most virtual reality, elderly assistance, home robotics, and other consumer-level scene understanding applications.

\vspace{2mm}
\noindent{\bf Scale.}  We believe that \OURS is the largest RGB-D dataset available.  The BuldingParser dataset \cite{armeni20163d} 
provides data for 270 rooms spanning 6,020$\text{m}^2$ of floor space.  ScanNet \cite{dai2017scannet}, provides images covering 78,595$\text{m}^2$ of surface area 
spanning 34,453$\text{m}^2$ of floor space in 707 distinct rooms.   Our dataset covers 219,399$\text{m}^2$ of surface area in 2056 rooms with 46,561$\text{m}^2$ of floor space.  This scale provides new opportunities for training data-hungry algorithms.

\section{Learning from the Data}
\label{sec:learning}

The following subsections describe several tasks leveraging these unique properties of the 
\OURS dataset to provide new ways to learn representations of scenes. For all experiments, we have split the dataset into 61 scenes for training, 11 for validation, and 18 for testing (see the supplemental materials for details).


\subsection{Keypoint Matching}

Matching keypoints to establish correspondences between image data is an important task for many applications including mapping, pose estimation, recognition, and tracking. 
With the recent success of neural networks, several works have begun to explore the use of deep learning techniques for training state-of-the-art keypoint descriptors that can facilitate robust matching between keypoints and their local image features \cite{yi2016lift,simo2015discriminative,han2015matchnet}. 
To enable training these deep descriptors, prior works leverage the vast amounts of correspondences found in existing RGB-D reconstruction datasets \cite{schmidt2017self,zeng20163dmatch}. 

\begin{figure}[t]
	\includegraphics[width=1\linewidth]{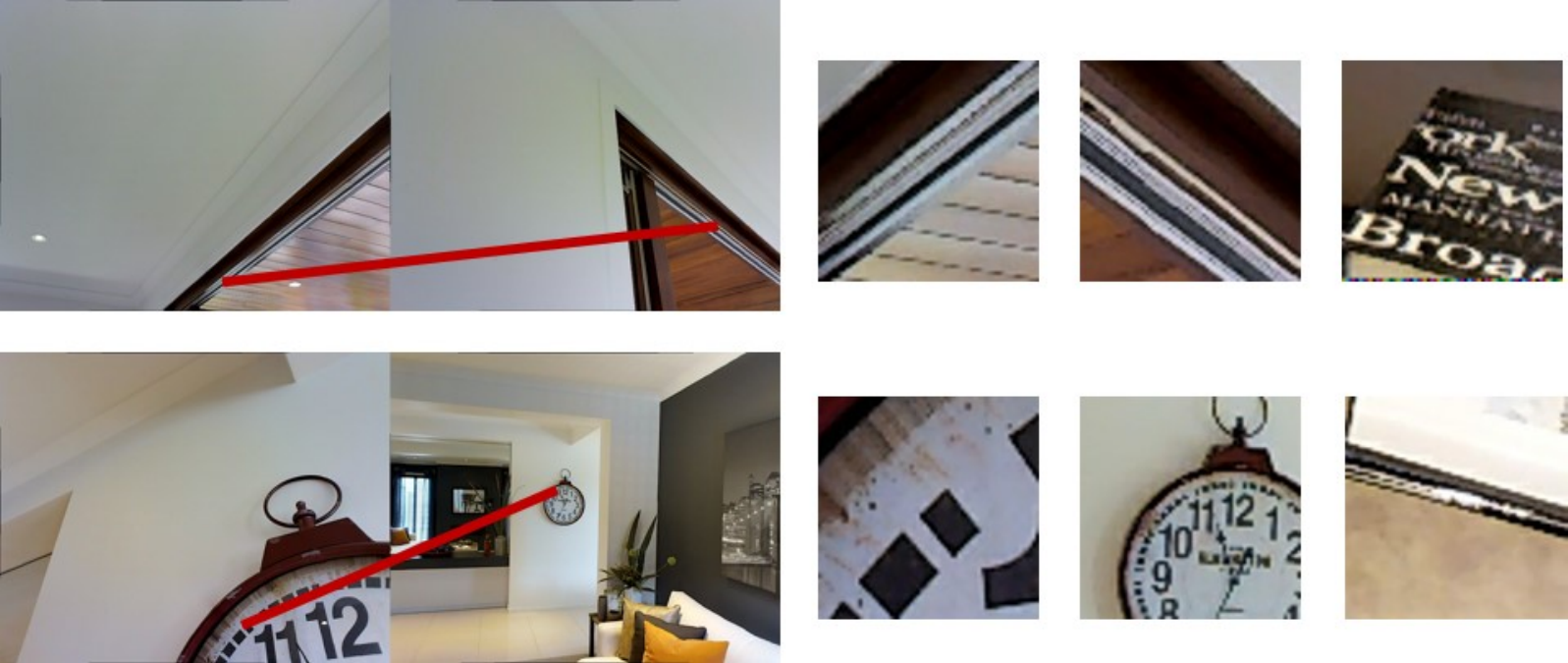}
	\caption{Example training correspondences (left) and image patches (right) extracted from \OURS. Triplets of matching patches (first and second columns) and non-matching patches (third column) are used to train our deep local keypoint descriptor.}
	\label{fig:keypointmatching}
\end{figure}

With the precise global alignment of RGB-D data and comprehensive view sampling, our \OURS dataset provides the unique opportunity to retrieve high quality, wide-baselined correspondences between image frames (see Figure \ref{fig:keypointmatching}). 
We demonstrate that by pretraining deep local descriptors over these correspondences, we can learn useful features to enable training even stronger descriptors. 
More specifically, we train a convolutional neural network (ResNet-50 \cite{he2016deep}) to map an input image patch to a 512 dimensional descriptor. 
Similar to state of the art by \cite{hoffer2016deep}, we train the ConvNet in a triplet Siamese fashion, where each training example contains two matching image patches and one non-matching image patch.
Matches are extracted from SIFT keypoint locations which project to within $0.02$m of each other in world space and have world normals within $100^\circ$. 
To supervise the triplet model, we train with an L2 hinge embedding loss. 

For evaluation, we train on correspondences from 61 \OURS scenes and 17 SUN3D scenes, and test on ground truth correspondences from 8 held out SUN3D scenes. The SUN3D ground truth correspondences and registrations are obtained from \cite{StructuredGlobalRegistration}, using the training and testing scenes split from \cite{zeng20163dmatch}.
As in \cite{han2015matchnet}, we measure keypoint matching performance with the false-positive rate (error) at 95\% recall, the lower the better.
We train three models - one trained on \OURS data only, one trained on SUN3D data only, and another pretrained on \OURS and fine-tuned on SUN3D.
Overall, we show that pretraining on \OURS yields a descriptor that achieves better keypoint matching performance on a SUN3D benchmark. 

\begin{table}[t]
	\setlength{\tabcolsep}{1.9pt}
	\centering\footnotesize
	\begin{tabular}{|c|c|} 
		\hline
		SURF & 46.8\% \\
		SIFT & 37.8\% \\
		ResNet-50 w/ \OURS & 10.6\% \\ 
		ResNet-50 w/ SUN3D & 10.5\% \\ 
		ResNet-50 w/ \OURS + SUN3D & 9.2\% \\
		\hline
	\end{tabular}
	\vspace{0.1cm}
	\caption{\textbf{Keypoint matching results.} Error (\%) at 95\% recall on ground truth correspondences from the SUN3D testing scenes. We see an improvement in performance from pretraining on \OURS.}
	\label{table:classification}
\end{table}


\subsection{View Overlap Prediction}
\label{sec:overlap_prediction}

Identifying previously visited scenes is a fundamental step for many reconstruction pipelines -- i.e., to detect loop closures. 
While previous RGB-D video datasets may only have few instances of loop closures, the \OURS dataset has a large number of view overlaps between image frames due to the panoramic nature and comprehensive viewpoint sampling of the capturing process. 
This large number of loop closures provides an opportunity to train a deep model to recognize loop closures, which can be incorporated in future SLAM reconstruction pipelines.

In this work, we formalize loop closure detection as an image retrieval task.  Given a query image, the goal is to find other images with ``as much overlap in surface visibility as possible.''  We quantify that notion 
as a real-numbered value modeled after intersection over union (IOU):
$ overlap(A,B) =min(\hat{A}, \hat{B}) / (|A| + |B| - min(\hat{A}, \hat{B})) $
where $A$ and $B$ are images, $|A|$ is the number of pixels with valid depths in $A$, 
$\hat{A}$ is the number of pixels of image $A$ whose projection into world space lie within 5cm 
of any pixel of $B$.

We train a convolutional neural network (ResNet-50 \cite{he2016deep}) to map each frame to features, 
where a closer L2 distance between two features indicates a higher overlap. 
Similar to keypoint matching, we train this model in a triplet Siamese fashion, using the distance ratio loss from \cite{hoffer2016deep}.
However, unlike the keypoint matching task, where there is a clear definition of "match" and "non-match," the overlap function can be any value ranging from 0 to 1. Therefore we add a regression loss on top of the triplet loss that directly regresses the overlap measurement between the "matching" image pairs (overlap ratio greater than 0.1).

\begin{figure}[t]
	\includegraphics[width=\linewidth]{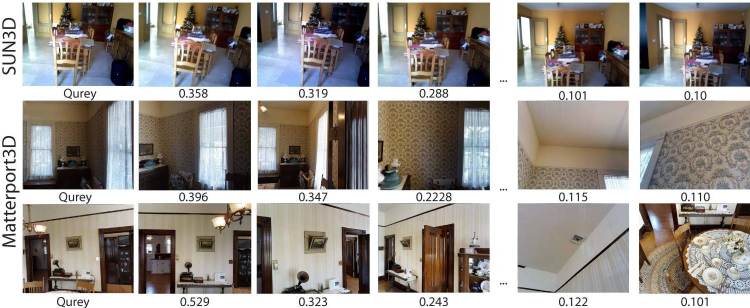}
	\caption{Example overlap views from SUN3D and \OURS ranked by their overlap ratio. In contrast to RGB-D video datasets captured with hand-held devices like SUN3D, \OURS provides a larger variety of camera view points and wide baseline correspondences, which enables training a stronger model for view overlap prediction under such challenging cases.\vspace{-0.2cm}}
	\label{fig:overlaps}
\end{figure}

\begin{table}[t]
	\setlength{\tabcolsep}{4pt}
	\centering\footnotesize
	\begin{tabular}{c | c |c |c} 
		\hline
		Training & Testing &triplet	&triplet + regression \\
		\hline
		\OURS&	     SUN3D&     74.41&	81.97\\
		SUN3D&	     SUN3D&     79.91&	83.34\\
		\OURS + SUN3D& SUN3D& 	84.10&	85.45\\
		\hline
		\OURS &  \OURS &	48.8&	53.6\\
		\hline
	\end{tabular}
	\vspace{2mm}
	\caption{\label{table:overlap_sun} \textbf{View overlap prediction results.} Results on SUN3D and \OURS dataset measured by normalized discounted cumulative gain. From the comparison we can clearly see the performance improvement from training data with \OURS and from adding the extra overlap regression loss. We also note that overlap prediction is much harder in the \OURS dataset due to the wide baselines between camera poses.}
\end{table}

\begin{figure*}
	\includegraphics[width=\linewidth]{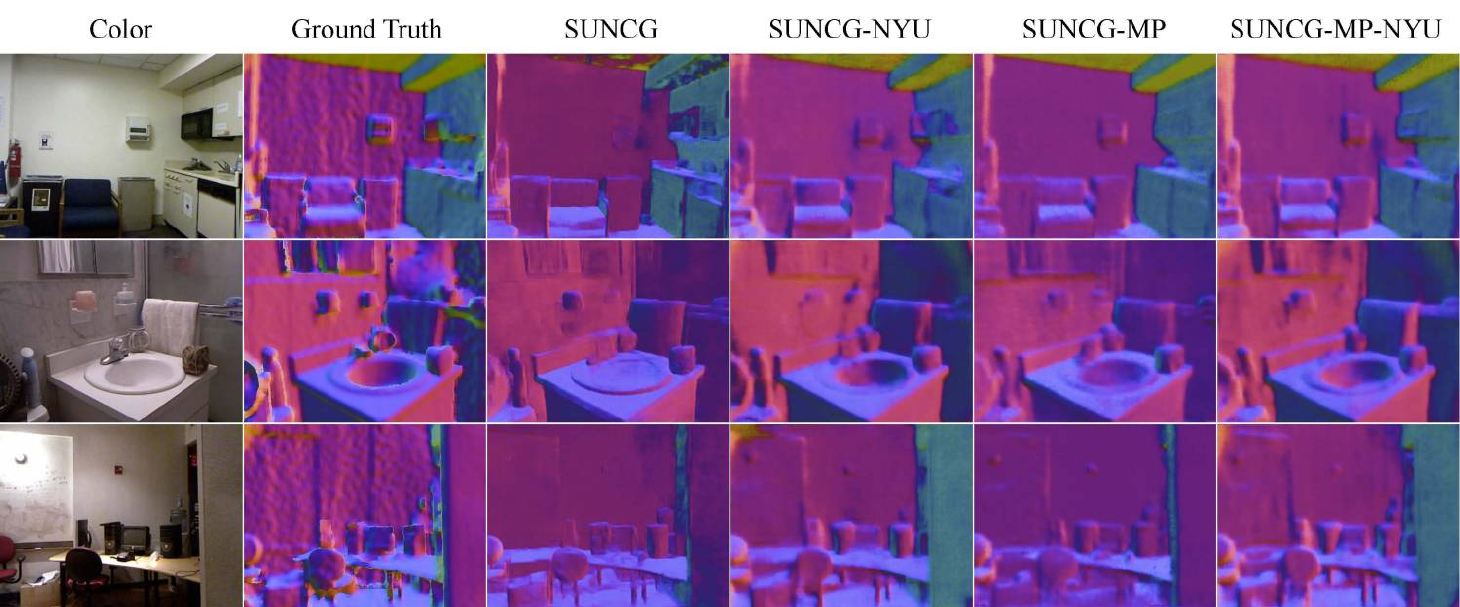}
	\caption{Examples of surface normal estimation. We show results of images from NYUv2 testing set. The results from the model fine-tuned on \OURS (SUNCG-MP) shows the best quality visually, as it starts to capture small details while still produces smooth planar area. The model further fine-tuned on NYUv2 (SUNCG-MP-NYU) achieves the best quantitatively performance but tends to produce comparatively noisy results. }
	\label{fig:normal_estimation}
\end{figure*}

Table \ref{table:overlap_sun}  shows an evaluation of this network trained on the \OURS training set and then tested on both the \OURS test set and the SUN3D dataset \cite{xiao2013sun3d}.  For each test, we generate a retrieval list sorted by predicted distance and evaluate it by computing the normalized discounted cumulative gain between predicted list and the best list from ground truth.
To mimic real reconstruction scenarios, we only consider candidate image pairs that have travel distance greater than 0.5m apart.
The experimental results show that training on the \OURS dataset helps find loop closures when testing on SUN3D, and that the extra supervision of overlap ratio regression helps to improve performance on both test sets. We can also notice that overlap prediction is much harder in our \OURS dataset due to the wide baseline between camera poses, which is very different from data captured with hand held devices like SUN3D (Figure \ref{fig:overlaps}).



\subsection{Surface Normal Estimation}
\label{sec:normal_estimation}

Estimating surface normals is a core task in scene reconstruction and scene understanding.   Given a color image, the task is to estimate the surface normal direction for each pixel.  Networks have been trained to perform that task using RGB-D datasets in the past \cite{eigen2015predicting,li2015depth,bansal2016marr,wang2015designing,zhang2016physically}.   However, the depths acquired from commodity RGB-D cameras are generally very noisy, and thus provide poor training data.  In contrast, the Matterport camera acquires depth continuously as it rotates for each panorama and synthesizes all the data into depth images aligned with their color counterparts which produces normals with less noise. 

\begin{table}[t]
	\center
	\scalebox{0.55}{
		\begin{tabular}{c c c | c c c c c}
			\hline
			Train Set 1 & Train Set 2 & Train Set 3 & Mean($^{\circ}$)$\downarrow$ & Median($^{\circ}$)$\downarrow$ & 11.25($\%$)$\uparrow$ & 22.5($\%$)$\uparrow$ & 30($\%$)$\uparrow$ \\
			\hline
			SUNCG & - & - & 28.18 & 21.75 & 26.45 & 51.34 & 62.92 \\
			SUNCG & NYUv2 & - & 22.07 & 14.79 & 39.61 & 65.63 & 75.25 \\
			\hline
			MP & - & - & 31.23 & 25.95 & 18.17 & 43.61 & 56.69\\
			MP & NYUv2 & - & 24.34 & 16.94 & 35.09 & 60.72 & 71.13 \\
			SUNCG & MP & - & 26.34 & 21.08 & 23.04 & 53.36 & 67.45 \\
			SUNCG & MP & NYUv2 & \bf 20.89 & \bf 13.79 & \bf 42.29 & \bf 67.82 & \bf 77.16 \\
			\hline
		\end{tabular}
	}
	\caption{{\bf Surface normal estimation results.} Impact of training with \OURS (MP) on performance in the NYUv2 dataset. The columns show the mean and median angular error on a per pixel level, as well as the percentage of pixels with error less than $11.25^{\circ}$, $22.5^{\circ}$, and $30^{\circ}$. }
	\label{tab:normal_prediction_main}
	\center
	\scalebox{0.7}{
		\begin{tabular}{c c | c c c c c}
			\hline
			Train & Test & Mean($^{\circ}$)$\downarrow$ & Median($^{\circ}$)$\downarrow$ & 11.25($\%$)$\uparrow$ & 22.5($\%$)$\uparrow$ & 30($\%$)$\uparrow$ \\
			\hline
			MP & NYUv2 & 26.34 & 21.08 & 23.04 & 53.35 & 67.45 \\
			NYUv2 & NYUv2 & 22.07 & 14.79 & 39.61 & 65.63 & 75.25 \\
			MP & MP & 19.11 & 10.44 & 52.33 & 72.22 & 79.46 \\ 
			NYUv2 & MP & 33.91 & 25.07 & 23.98 & 46.26 & 56.45 \\
			\hline
		\end{tabular}
	}
	\caption{{\bf Surface normal estimation cross dataset validation}. We investigate the influence of training and testing the model using permutation of datasets. Notice how the \OURS dataset is able to perform well on NYUv2, while the converse is not true. }
	\label{tab:normal_prediction_cross}
\end{table}

In this section, we consider whether the normals in the \OURS dataset can be used to train better models for normal prediction on other datasets.  For our study, we use the model proposed in Zhang \etal \cite{zhang2016physically}, which achieves the state of the art performance on the NYUv2 dataset.
The model is a fully convolutional neural network consisting of an encoder, which shares the same architecture as VGG-16 from the beginning till the first fully connected layer, and a purely symmetric decoder.
The network also contains short-cut link to copy the high resolution feature from the encoder to the decoder to bring in details, and forces the up pooling to use the same sampling mask from the corresponding max pooling layer.

Zhang \etal \cite{zhang2016physically} demonstrate that by pretraining on a huge repository of high-quality synthetic data rendered from SUNCG \cite{song2016semantic} and then fine-tuning on NYUv2, the network can achieve significantly better performance than directly training on NYUv2.
They also point out that the noisy ground truth on NYUv2 provides inaccurate supervision during the training, yielding results which tend to be blurry. With an absence of real-world high-quality depths, their model focuses solely on the improvement from
pretraining on synthetic scenes and fine-tuning on real scenes.  

We use \OURS data as a large-scale real dataset with high-quality surface normal maps for pretraining, and train the model with a variety of training strategies.
For the \OURS data, we use only the horizontal and downward looking views as they are closer to canonical views a human observer would choose to look at a scene.
Table~\ref{tab:normal_prediction_main} shows the performance of surface normal estimation. As can be seen, the model pretrained using both the synthetic data and \OURS data (the last row) outperforms the one using only the synthetic data (the 2nd row) and achieves best performance.

We show the cross dataset accuracy in Table~\ref{tab:normal_prediction_cross}. 
We train models by first pretraining on synthetic data and then fine-tuning on each dataset; i.e., NYUv2 and \OURS, respectively.
We evaluate two models on the test set of each dataset.
The model trained on each dataset provides the best performance when testing on the same dataset. 
However, the NYUv2 model performs poorly when testing on \OURS, while the \OURS model still performs reasonably well on NYUv2.
This demonstrates that model trained on \OURS data generalizes much better, with its higher quality of depth data and diversity of viewpoints.

Figure ~\ref{fig:normal_estimation} shows results on NYUv2 dataset. Compared to the model only trained on the synthetic data (SUNCG) or NYUv2 (SUNCG-NYU), the model fine-tuned on \OURS shows the best visual quality, as it captures more detail on small objects, such as the paper tower and fire alarm on the wall, while still producing smooth planar regions. 
This improvement on surface normal estimation demonstrates the importance of having high quality depth.
The model further fine-tuned on NYUv2 (SUNCG-MP-NYU) achieves the best quantitatively performance, but tends to produce comparatively noisy results since the model is ``contaminated'' by the noisy ground truth from NYUv2.

\subsection{Region-Type Classification}
\label{sec:region_classification}

Scene categorization is often considered as the first step for high-level scene understanding and reasoning. 
With the proposed dataset, which contains a large variety of indoor environments, we focus our problem on indoor region (room) classification -- given an image, classify the image based on the semantic category of the region that contains its viewpoint (e.g., the camera is in a bedroom, or the camera is in a hallway).

Unlike the semantic voxel labeling problem, region-level classification requires understanding global context that often goes beyond single view observations. 
While most of the scene categorization datasets \cite{xiao2010sun,zhou2014learning} focus on single view scene classification, this dataset provides a unique opportunity to study the relationship between image field of view and scene classification performance.

As ground truth for this task, we use the 3D region annotations provided by people as described in Section \ref{sec:semantic_annotation}. We choose the 12 most common categories in the dataset for this experiment. We assign the category label for each panorama or single image according to the label provided for the region containing it.  We then train a convolutional neural network (ResNet-50 \cite{he2016deep}) to classify each input image to predict the region type.

Table \ref{table:roomtype} shows the classification accuracy (number of true positives over the total number of instances per region type). By comparing the accuracy between [single] and [pano], we can see an improvement in performance from increased image field of view for most region types. The lower performance in lounge and family room is due to confusion with other adjacent regions (e.g. they are often confused with adjacent hallways and kitchens, which are more visible with wider fields of view).

\begin{table*}
	\setlength{\tabcolsep}{6pt}
	\footnotesize
    \centering
	\begin{tabular}{c|ccccccccccccccccccc}
		\hline 
		class & office & lounge & familyroom & entryway & dining room & living room& stairs& kitchen  &porch & bathroom &  bedroom &hallway \tabularnewline
		\hline 
		single & 20.3&21.7&16.7& 1.8&20.4&27.6&49.5&52.1&57.4&44.0&43.7&44.7\tabularnewline
		pano&  26.5&15.4&11.4 &3.1&27.7&34.0&60.6&55.6&62.7&65.4&62.9&66.6 \tabularnewline
		\hline
	\end{tabular}
	\caption{\label{table:roomtype}{\bf Region-type classification results.} Each entry lists the prediction accuracy (percentage correct).  By comparing the accuracy between [single] and [pano] we can see an improvement from increased image field of view for most regiontypes. However, the lower performance on lounge and family room may be caused by confusion from seeing multiple rooms in one panorama.}
\end{table*}

\subsection{Semantic Voxel Labeling}
\label{sec:semantic_voxel_labeling}
Semantic voxel labeling -- i.e., predicting a semantic object label for each voxel -- is a fundamental task for semantic scene understanding; it is the analog of image segmentation in 3D space.
We follow the description of the semantic voxel labeling task as introduced in ScanNet~\cite{dai2017scannet}.

For training data generation, we first voxelize the training scenes into a dense voxel grid of $2$cm$^3$ voxels, where each voxel is associated with its occupancy and class label, using the object class annotations.
We then randomly extract subvolumes from the scene of size $1.5$m $\times$ $1.5$m $\times$ $3$m ($31 \times 31 \times 62$ voxels). Subvolumes are rejected if $< 2 \%$ of the voxels are occupied or $< 70 \%$ of these occupied voxels have valid annotations. Each subvolume is up-aligned, and augmented with $8$ rotations.

\begin{figure}[t]
	\includegraphics[width=\linewidth]{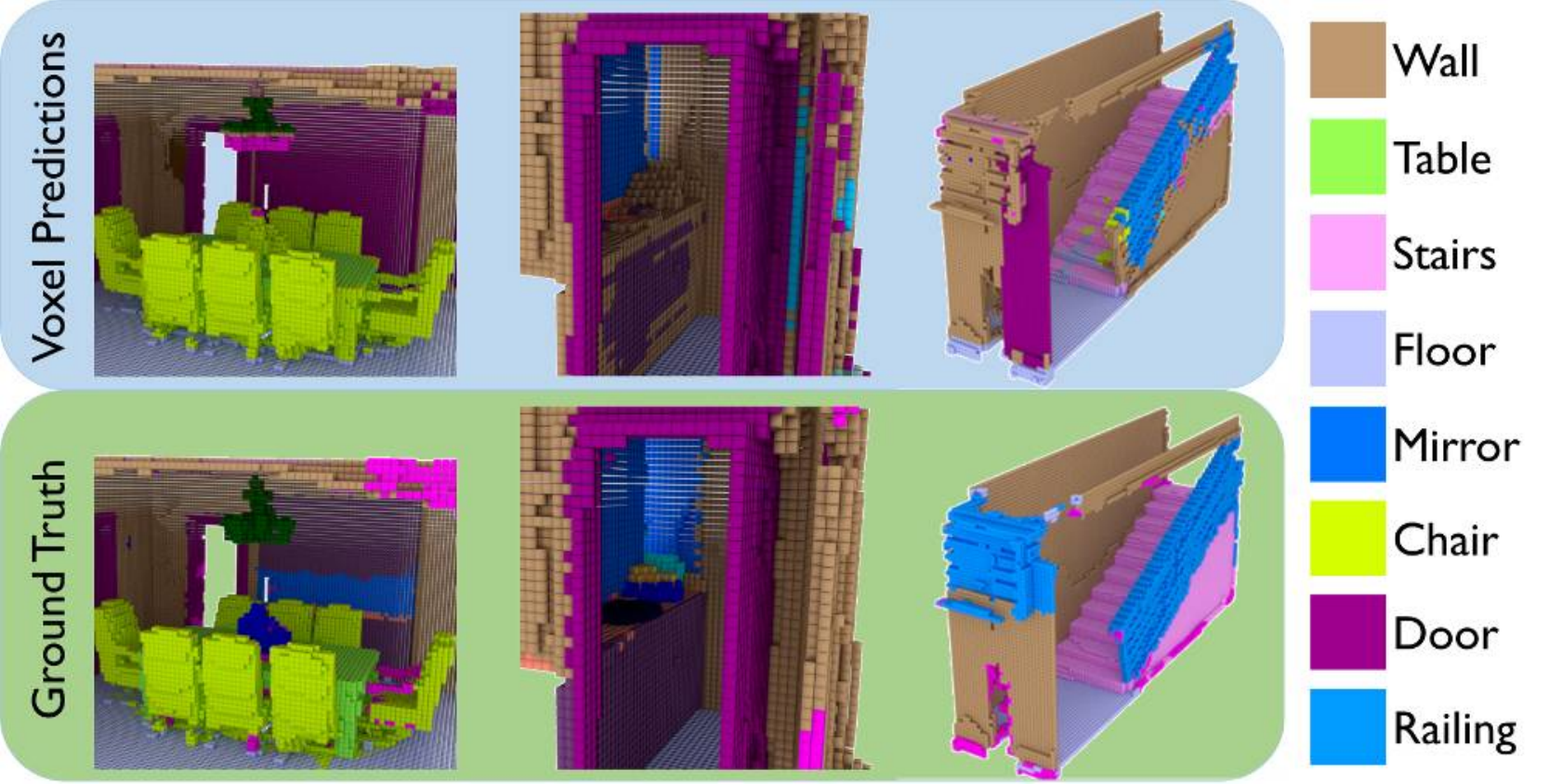}
	\caption{Semantic voxel labeling results on our \OURS test scenes.\vspace{-0.4cm}}
	\label{fig:sem_vox_label}
\end{figure}

We use 20 object class labels, and a network following the architecture of ScanNet~\cite{dai2017scannet}, and training with 52,355 subvolume samples (418,840 augmented samples).
Table~\ref{table:sem_vox_label} shows classification accuracy for our semantic voxel labeling on \OURS test scenes, with several visual results show in Figure~\ref{fig:sem_vox_label}.


\begin{table}
	\centering\footnotesize
	\begin{tabular}{|c|c|c|}
		\hline
		Class & \% of Test Scenes & Accuracy\\ \hline
		Wall & 28.9\% & 78.8\% \\ 
		Floor & 22.6\% & 92.6\% \\ 
		Chair & 2.7\% & 91.1\% \\ 
		Door & 5.0\% & 60.6\% \\ 
		Table & 1.7\% & 20.7\% \\ 
		Picture & 1.1\% & 28.4\% \\ 
		Cabinet & 2.9\% & 14.4\% \\
		Window & 2.2\% & 14.7\% \\ 
		Sofa & 0.1\% & 0.004\% \\ 
		Bed & 0.9\% & 1.0\% \\ 
		Plant & 2.0\% & 7.5\% \\ 
		Sink & 0.2\% & 23.8\% \\ 
		Stairs & 1.5\% & 54.0\% \\ 
		Ceiling & 8.1\% & 85.4\% \\ 
		Toilet & 0.1\% & 6.8\% \\ 
		Mirror & 0.4\% & 20.2\% \\ 
		Bathtub & 0.2\% & 5.1\% \\ 
		Counter & 0.4\% & 27.5\% \\ 
		Railing & 0.7\% & 18.3\% \\ 
		Shelving & 1.2\% & 16.6\% \\ 
		\hline
		Total & - & 70.3\% \\ \hline
	\end{tabular}
	\vspace{0.1cm}
	\caption{Semantic voxel label prediction accuracy on our \OURS test scenes.\vspace{-0.2cm}}
	\label{table:sem_vox_label}
\end{table}

\section{Conclusion}

We introduce \OURS, a large RGB-D dataset of \SCENECOUNT building-scale scenes.
We provide instance-level semantic segmentations on the full 3D reconstruction of each building.
In combination with the unique data characteristics of diverse, panoramic RGB-D views, precise global alignment over a building scale, and comprehensive semantic context over a variety of indoor living spaces, \OURS enables myriad computer vision tasks. 
We demonstrate that \OURS data can be used to achieve state of the art performance on several scene understanding tasks and release the dataset for research use.

\section{Acknowledgements}
The Matterport3D dataset is captured and gifted by Matterport for use by the academic community. 
We would like to thank Matt Bell, Craig Reynolds, and Kyle Simek for their help in accessing and processing the data, as well as the Matterport photographers who agreed to have their data be a part of this dataset.
Development of tools for processing the data were supported by Google Tango, Intel, Facebook, NSF
(IIS-1251217 and VEC 1539014/1539099), and a Stanford Graduate fellowship.

%% file: 0supp_body.tex




\section{Learning from the Data}

\begin{figure*}[p]
	\centering
	\includegraphics[width=\linewidth]{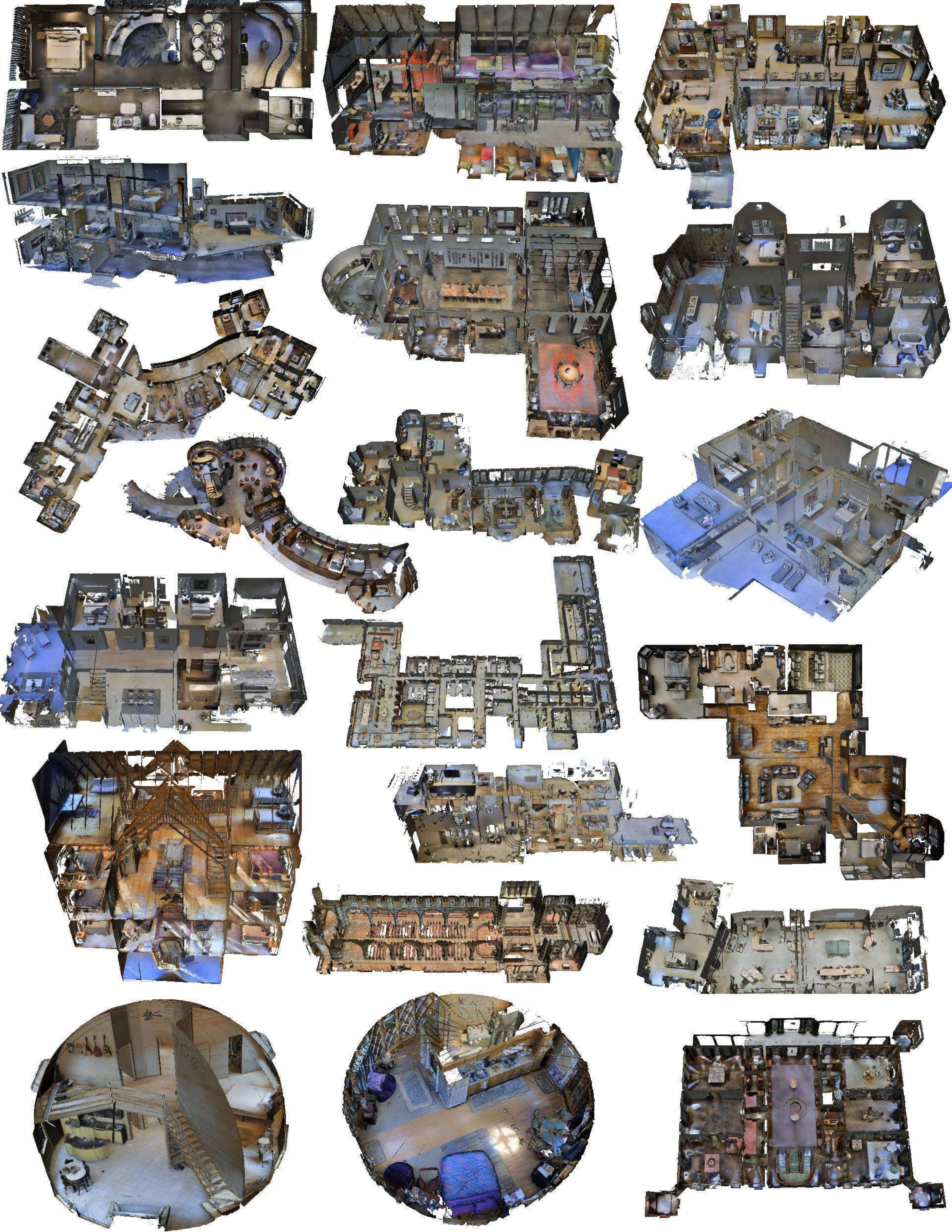} 
	\caption{{\bf Training Set.} Examples 1 - 20  }
	\label{fig:train_0}
\end{figure*}

\begin{figure*}[p]
	\centering
	\includegraphics[width=\linewidth]{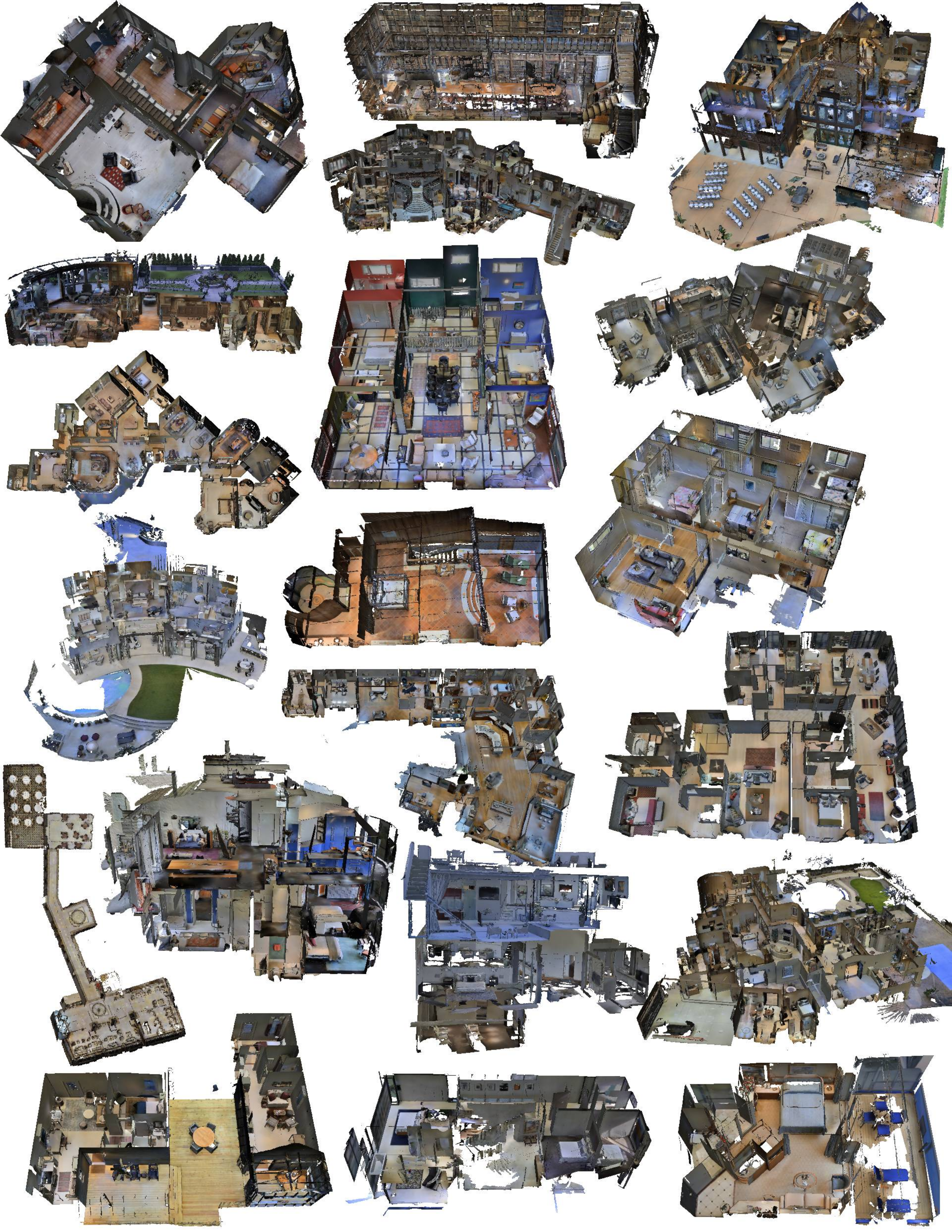} 
	\caption{{\bf Training Set.} Examples 21 - 40 }
	\label{fig:train_1}
\end{figure*}

\begin{figure*}[p]
	\centering
	\includegraphics[width=\linewidth]{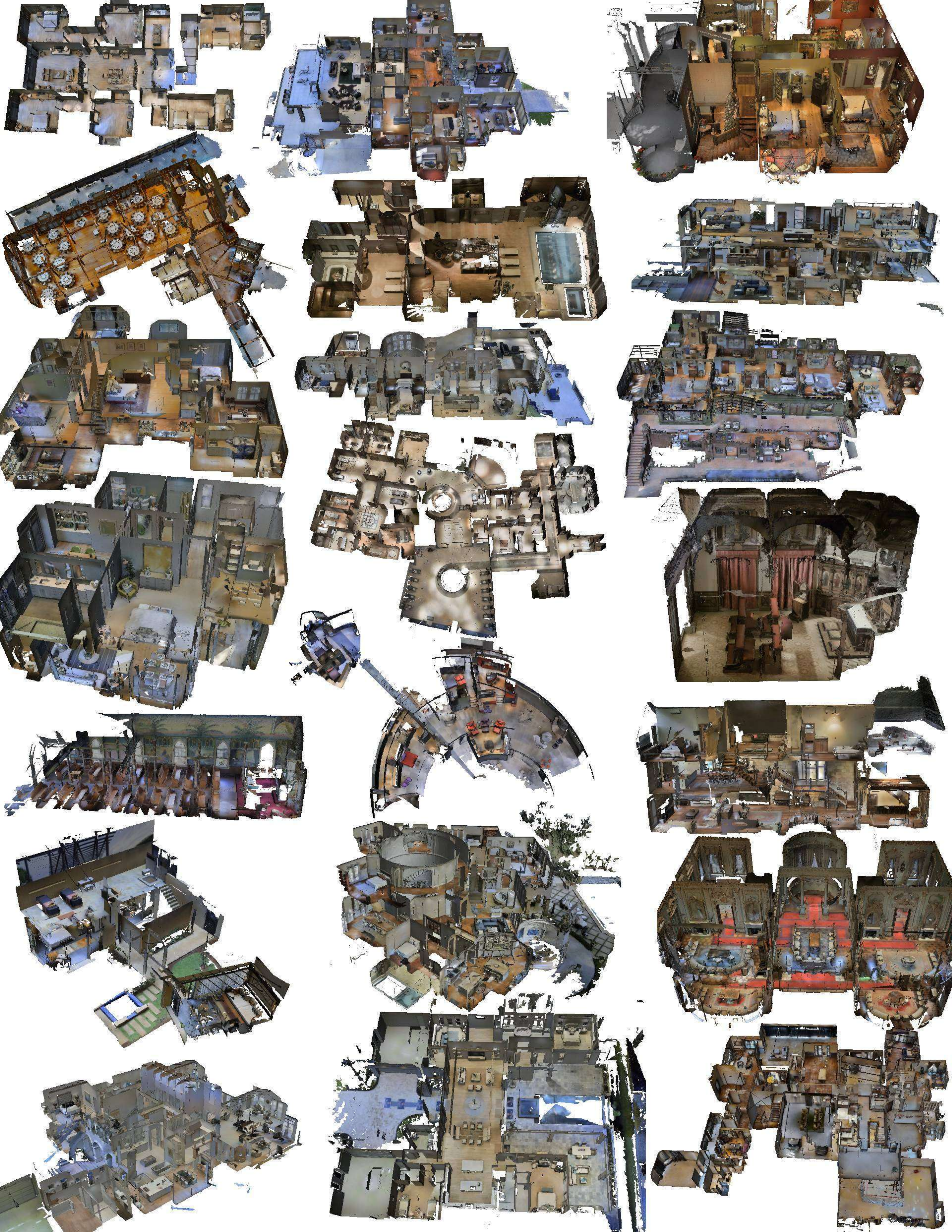} 
	\caption{{\bf Training Set.} Examples 41-61 }
	\label{fig:train_2}
\end{figure*}

\begin{figure*}[b]
	\centering
	\includegraphics[width=\linewidth]{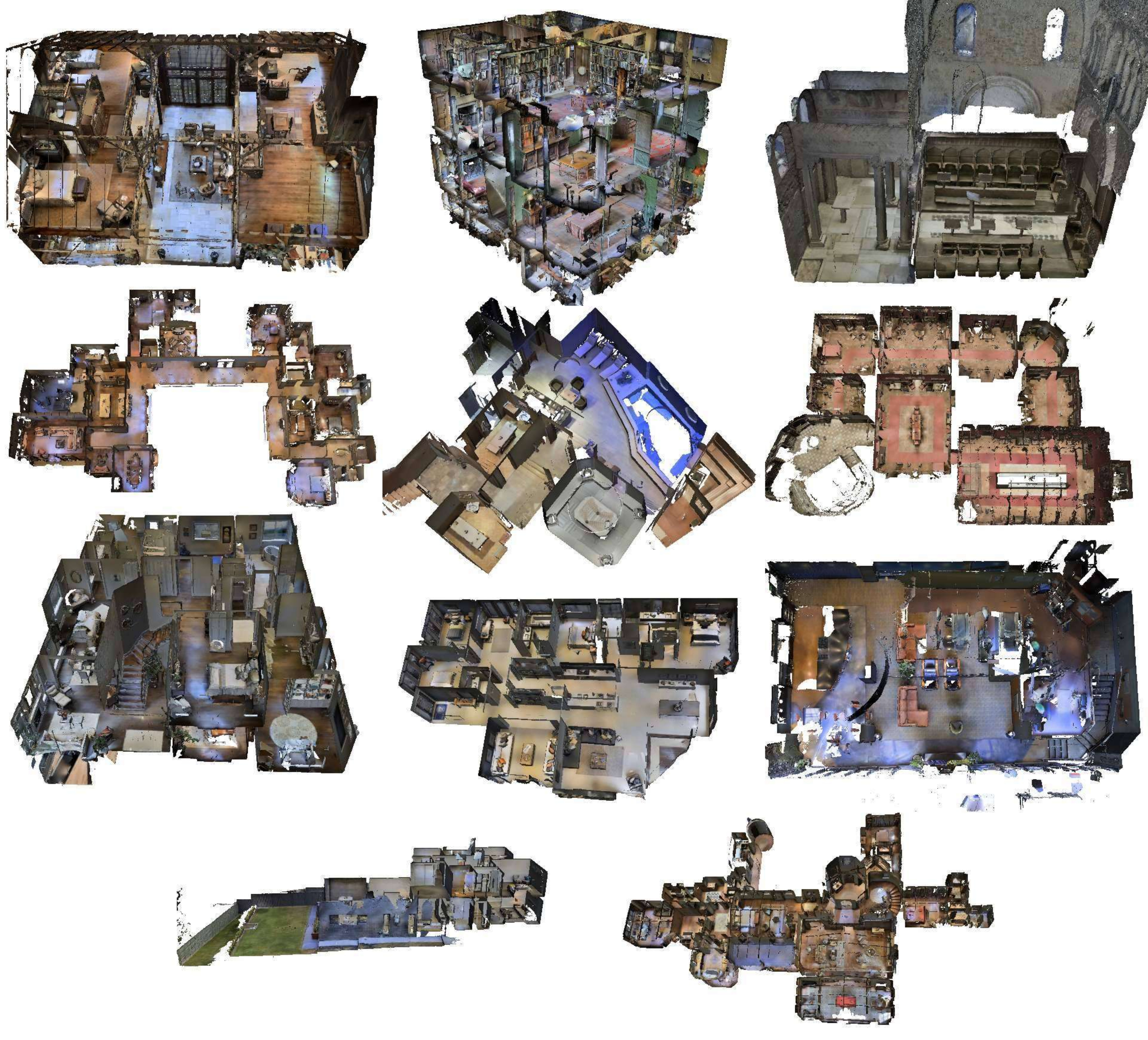} 
	\caption{\bf Validation Set. }
	\label{fig:validation_0}
\end{figure*}

\begin{figure*}[p]
	\centering
	\includegraphics[width=\linewidth]{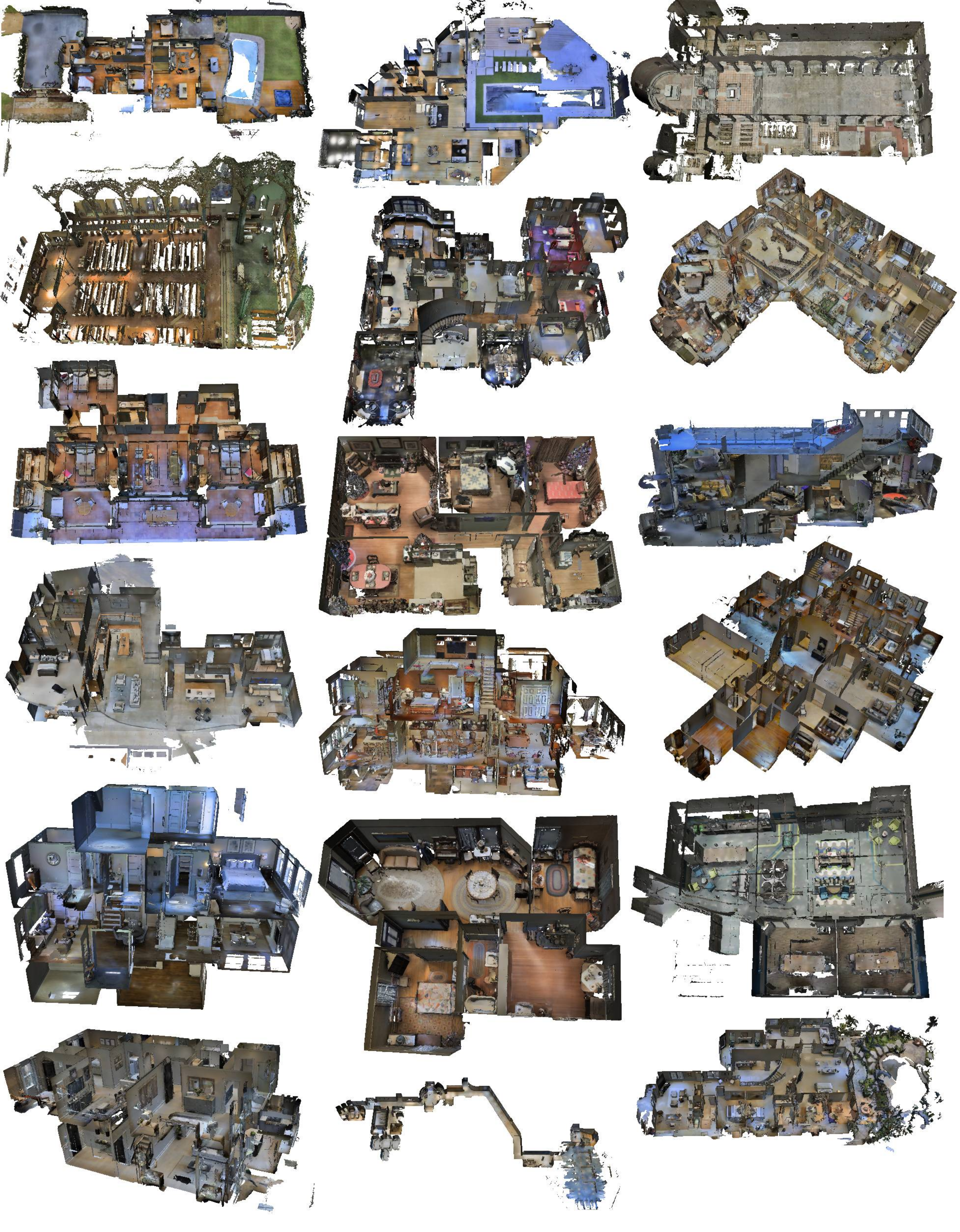} 
	\caption{\bf Test Set. }
	\label{fig:test_0}
\end{figure*}

The dataset is split into training, validation, and test set as shown in Fig.~\ref{fig:train_0} - \ref{fig:test_0}.
Each image shows the textured mesh for one scene from a bird's eye view.
These images are helpful for getting a sense of the diversity and scale of the scenes in the dataset.

\subsection{Keypoint Matching}

\begin{figure}[ht]
	\centering
	\includegraphics[width=\linewidth]{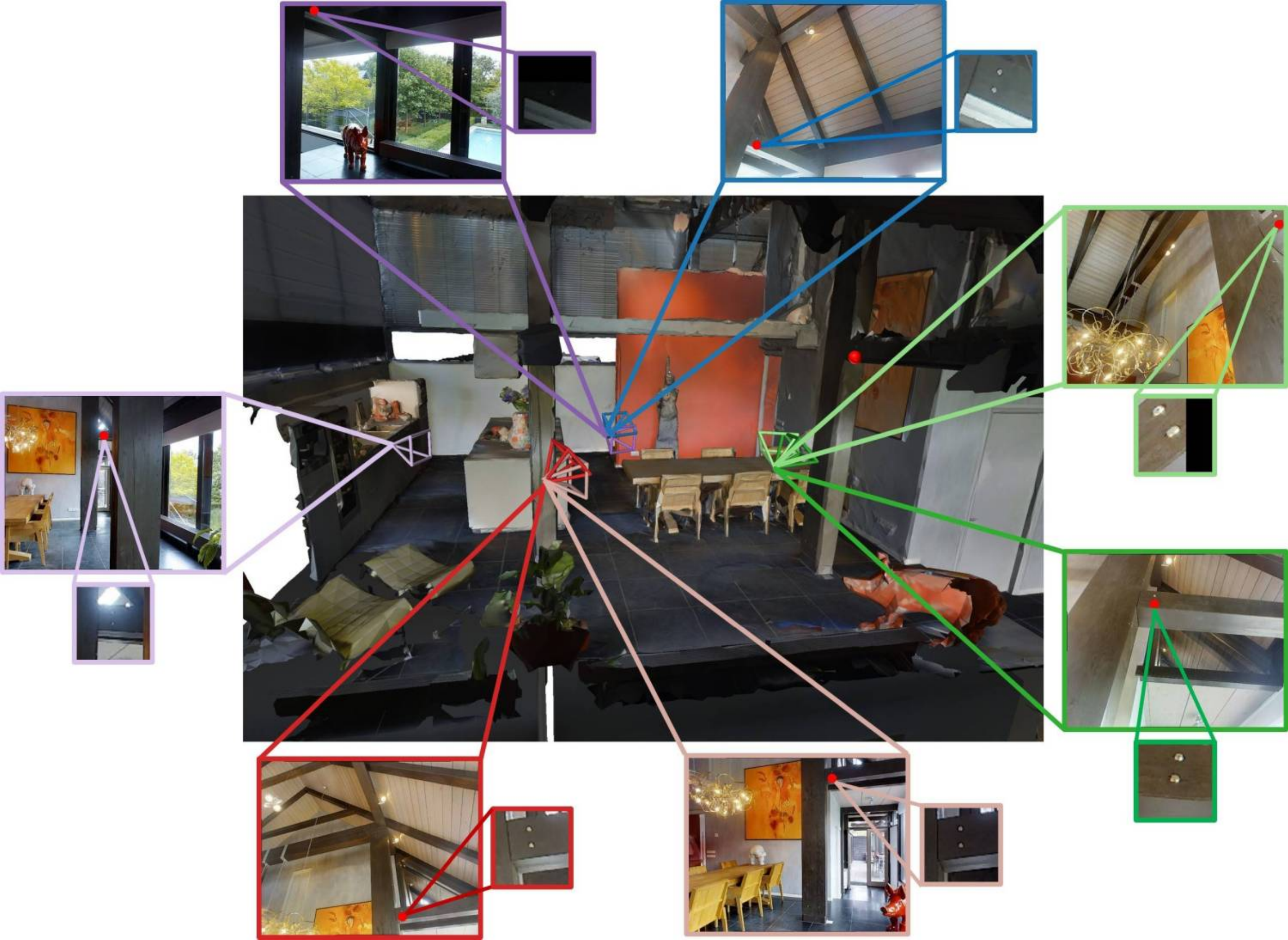}
	\caption{Visualization of the set of images visible to a keypoint. Each camera view sees the same key point (marked as red dots) from different view points; the smaller images are the patches that we use to train our keypoint descriptor.}
	\label{fig:keypoint_images}
\end{figure}

The first task considered in the paper is using \OURS to learn a descriptor for keypoint matching.  Unlike previous RGB-D datasets which contain video capture from hand-held sensors, \OURS comprises data captured from stationary cameras which comprehensively sample the viewpoint space. As shown in Fig~\ref{fig:keypoint_images}, this allows keypoints to be seen from a wide variety of differing views. Thus the \OURS data allows training for such scenarios which often provide significant challenges in keypoint matching and tracking (e.g., detecting loop closures).

\begin{figure*}[t]
	\centering
	\includegraphics[width=\linewidth]{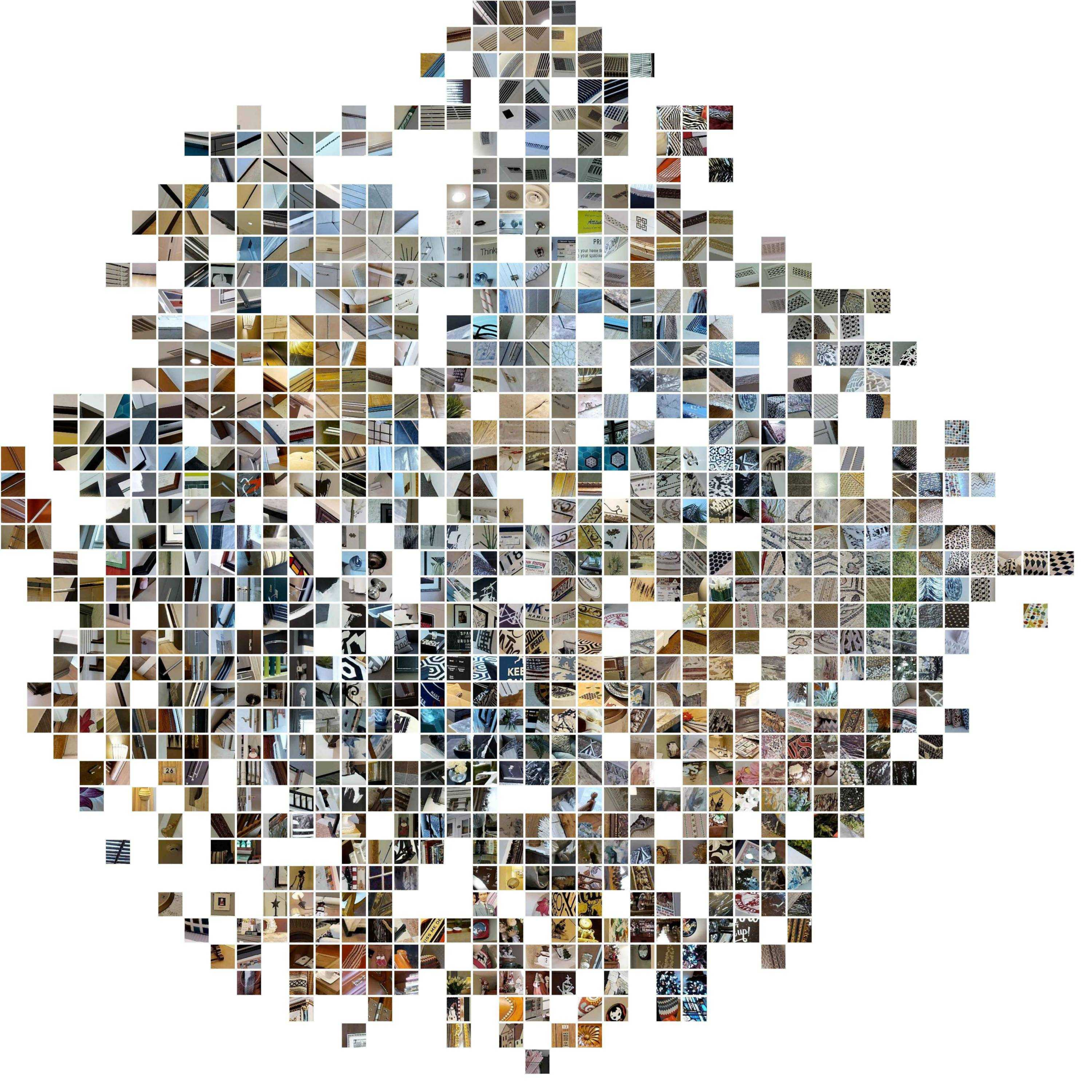}
	\caption{t-SNE embedding of descriptors from our triplet Siamese network trained for keypoint matching on the \OURS test set.}
	\label{fig:keypoint_tsne}
\end{figure*}

Fig.~\ref{fig:keypoint_tsne} shows a t-SNE embedding of local keypoint patches based on the descriptors from our triplet Siamese network, demonstrating the ability to cluster similar keypoints even with significant changes in view.

\subsection{View Overlap Prediction}

The second task investigated in the paper is predicting overlaps between pairs of views in the same scene -- i.e., the fraction pixels observing the same surface in both images.

\begin{figure*}[t]
	\centering
	\includegraphics[width=\linewidth]{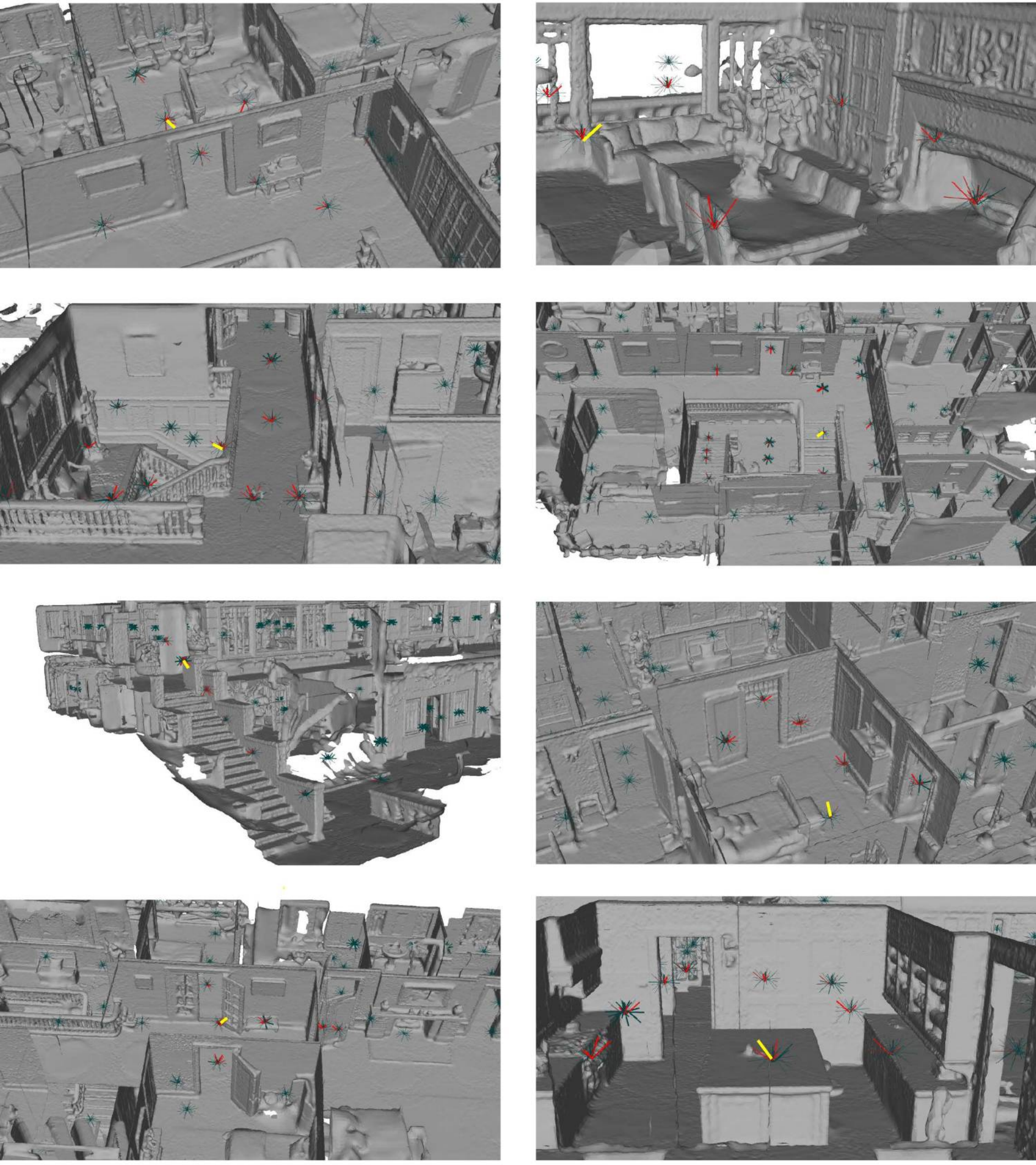}
	\caption{{\bf View Overlap Prediction.}: Eight views within the same scene showing overlaps between selected cameras (yellow) and all other cameras.  The colors ranging from thin dull cyan (no overlap) to thick bright red ($\geq$20\% overlap) indicate the fraction of overlap with the selected view.}
	\label{fig:overlap}
\end{figure*}

The \OURS dataset provides a unique opportunity to explore this task since it provides a large number and wide variety of overlapping views.
Fig.~\ref{fig:overlap} shows eight zoomed views of one scene.
In each image, the surface reconstruction is shown in shaded gray and cameras are shown as short line segments indicating view positions and directions.
The set of cameras within the same panorama look like a star because 18 separate images with 6 rotations and 3 tilt angles are captured for each tripod location.
Please note the density and diversity of views in the scenes. 

In each image of Fig.~\ref{fig:overlap}, a ``selected'' camera is highlighted in yellow, and all other cameras are colored according to how much they overlap with it -- thin dull cyan is 0\% overlap, thick bright red is $\geq$20\% overlap, and the line width and red channel scale linearly in between.   Please note that there are around a dozen significant overlaps in most of these examples, including overlaps between cameras with significantly different view directions and between viewpoints in different regions (please zoom the document as needed to see the figure at full resolution).  Predicting overlaps in these cases is a challenging task.

\subsection{Surface Normal Estimation}

\begin{figure*}[t]
	\centering
	\includegraphics[width=\linewidth]{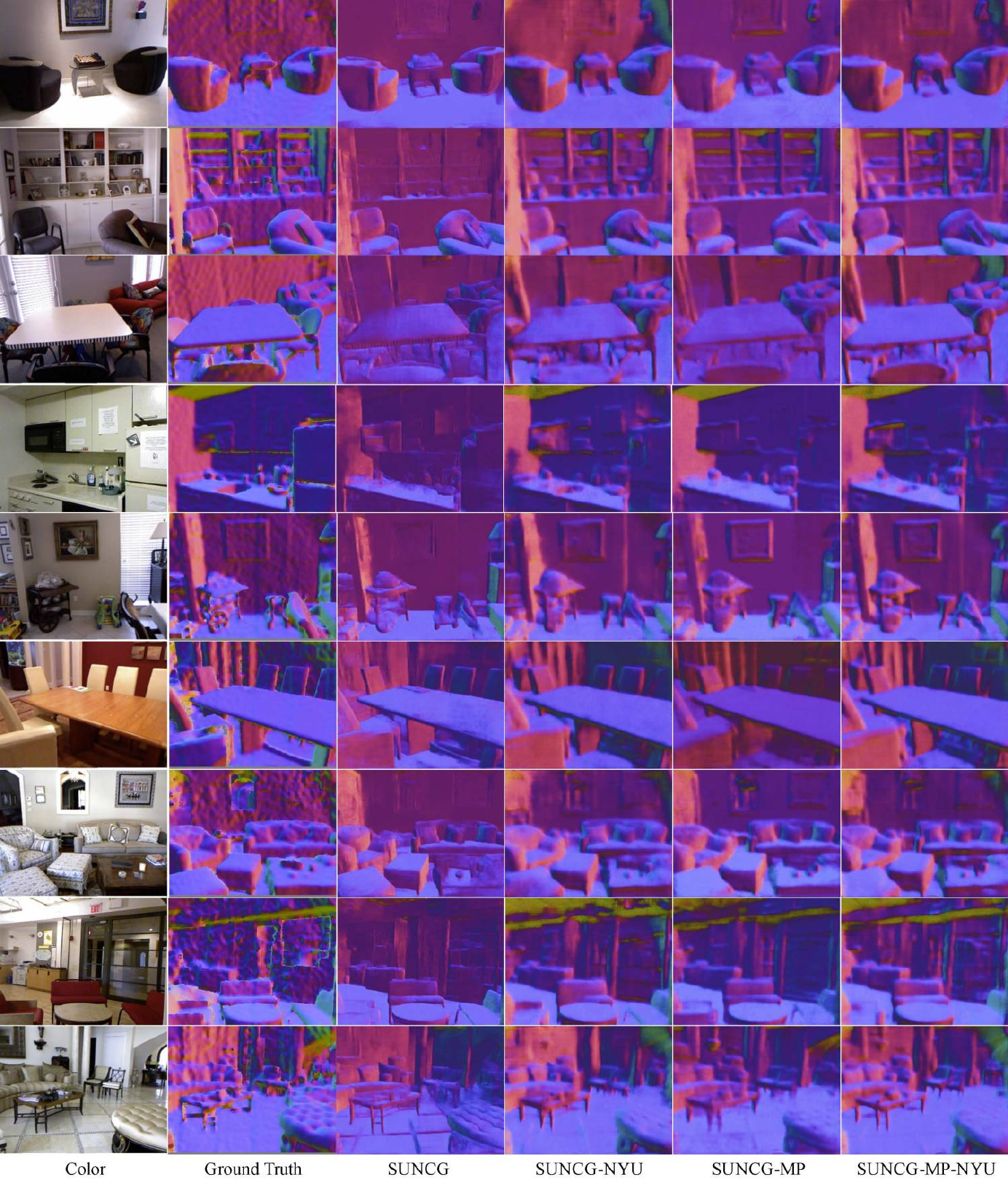}
	\caption{{\bf Surface Normal Estimation: Comparison of multiple training schema.} We compare the model pretrained with different datasets on the NYUv2 testing set. The 1st and 2nd columns show input color images and the ground truth normal map. The 3rd column shows the result of the model trained on physically based rendering. The 4th and 5th columns show results of models further finetuned on NYUv2 and \OURS. The last column shows results of models pretrained with both synthetic data and \OURS and then finetuned on NYUv2.}
	\label{fig:normal_comp}
\end{figure*}

The third task is to train a model to predict normals from RGB images.   Fig.~\ref{fig:normal_comp} shows comparisons of models trained with different combinations of datasets (please refer to paper for the details of each dataset). The 1st and 2nd columns show input color images and the ground truth normal map. The 3rd column shows the result of the model trained on physically based rendering from Zhang \etal \cite{zhang2016physically}. The 4th and 5th columns show results of models further finetuned on NYUv2 and \OURS. The last column shows results of models pretrained with both synthetic data and \OURS and then finetuned on NYUv2.

We can see that the ``SUNCG-MP'' model often produces clean result (e.g., in large areas of planar surface) with more details (e.g., painting on the wall, book on the table). 
Further finetuning on the noisy ground truth from NYUv2 actually hurts these desirable properties.

\begin{figure*}[t]
	\centering
	\includegraphics[width=\linewidth]{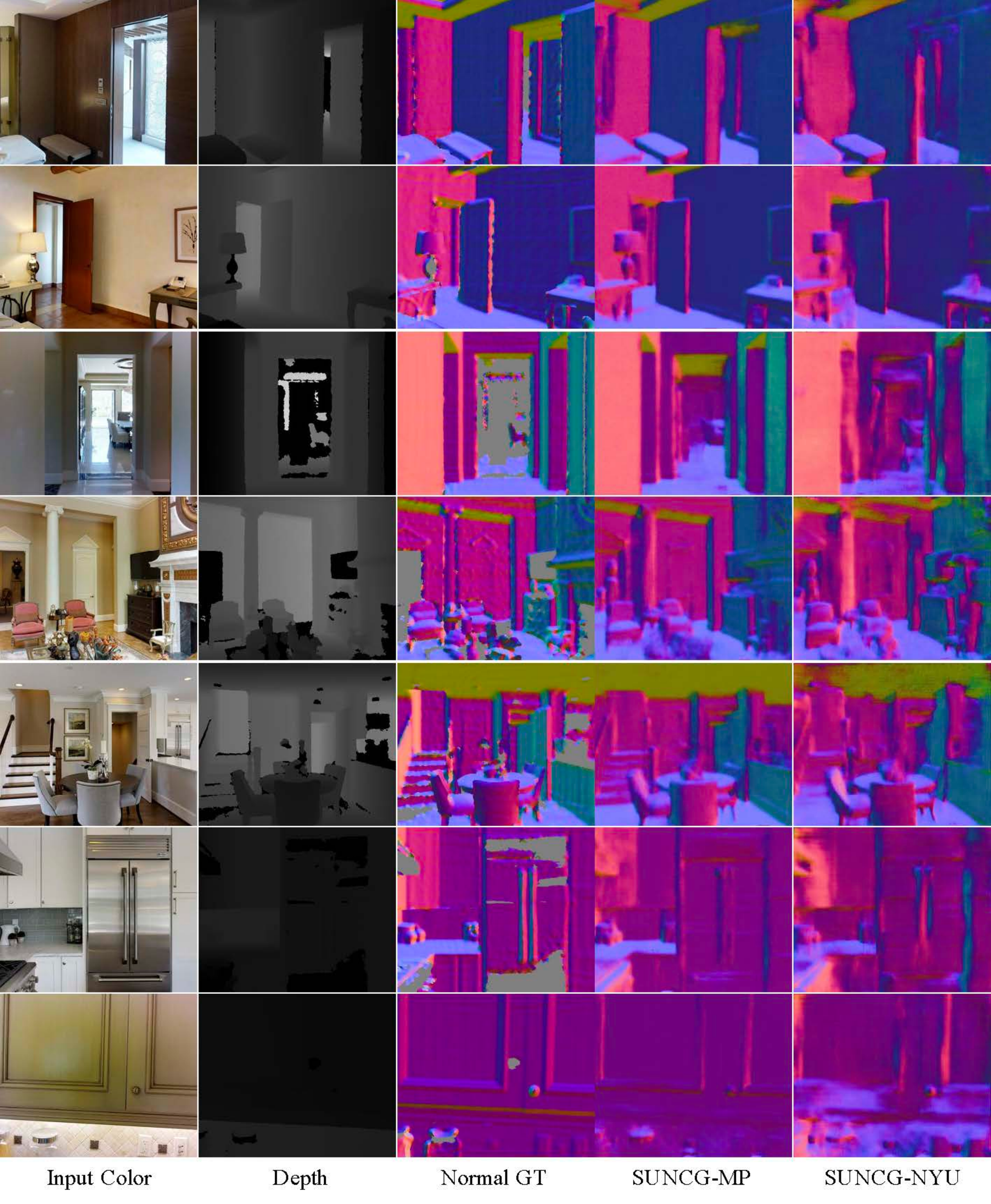}
	\caption{{\bf Surface Normal Estimation: Evaluation on \OURS images.} We evaluate two models trained on NYUv2 and \OURS respectively (both by finetuning the pretrained model on synthetic data) using images from \OURS. The quality of depth (2nd column) and surface normal (3rd column) is much better than that of the NYUv2 shown in Fig.~\ref{fig:normal_nyu}. The model trained on \OURS (4th column) does good job in predicting the surface normal, whereas model trained on NYUv2 (5th column) performs significantly worse.}
	\label{fig:normal_mat}
\end{figure*}

\begin{figure*}[t]
	\centering
	\includegraphics[width=\linewidth]{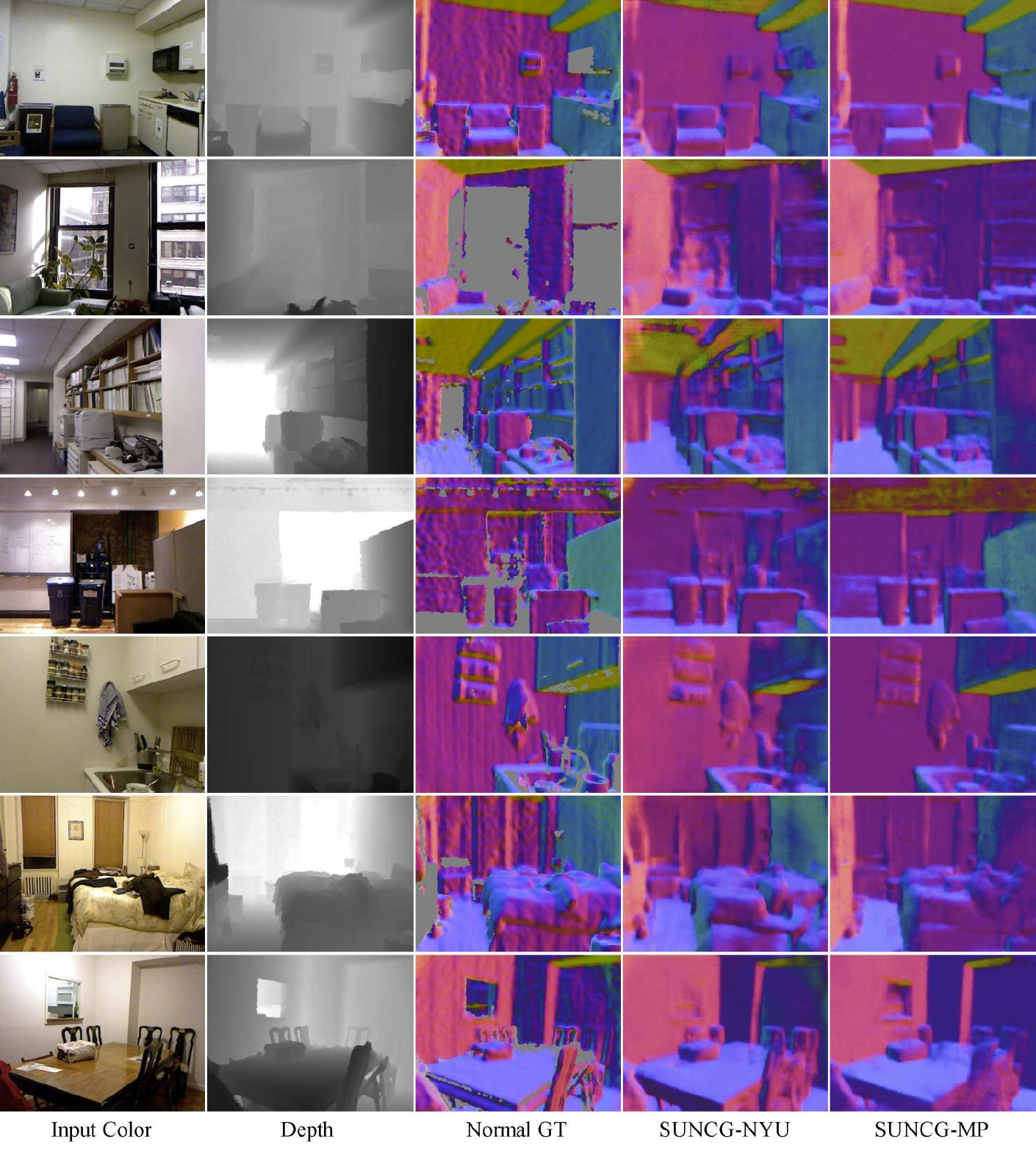}
	\caption{{\bf Surface Normal Estimation: Evaluation on NYUv2 images.} We evaluate two models trained on NYUv2 and \OURS respectively (both by finetuning the pretrained model on synthetic data) using images from NYUv2. The quality of depth (2nd column) and surface normal (3rd column) is much worse than that of the \OURS shown in Fig.~\ref{fig:normal_mat}. The model trained on NYUv2 (4th column) does good job in predicting the surface normal, while model trained on \OURS (5th column) still produces reasonably good results, sometimes even cleaner.}
	\label{fig:normal_nyu}
\end{figure*}

To further compare the quality of depth and surface normal provided in \OURS and NYUv2 and also their impacts on the model, we take the model pretrained on the synthetic data, and finetune on \OURS and NYUv2 respectively and evaluate both models on the testing set of two datasets.
Fig.~\ref{fig:normal_mat} and Fig.~\ref{fig:normal_nyu} show that qualitative results on images from \OURS and NYUv2 respectively.
We can see that:
\begin{itemize}
	
	\item The quality of depth and surface normal from \OURS (Fig.~\ref{fig:normal_mat} 2nd, 3rd column) is better than NYUv2 (Fig.~\ref{fig:normal_nyu} 2nd, 3rd column). The depth from \OURS presents cleaner flat surface with more details, whereas the noise in depth images from NYUv2 almost overwhelm local details.
	
	\item On images from \OURS, the model finetuned on \OURS (Fig.~\ref{fig:normal_mat} 4th column) produces good results containing both clean flat region and local details. However, the model trained on NYUv2 (Fig.~\ref{fig:normal_mat} 5th column) does not work well.
	
	\item On images from NYUv2, the model finetuned on NYUv2 (Fig.~\ref{fig:normal_nyu} 4th column) produces good results. The model trained on \OURS (Fig.~\ref{fig:normal_nyu} 5th column) still managed to produce reasonably good results, sometimes even better.
	
	\item Our model can predict correctly surface normal for those areas with missing depth (gray area in the surface normal map), which implies the potential of improving raw depth images from sensor using our model.
	
\end{itemize}

\subsection{Region-Type Classification}

\begin{figure*}[p]
	\centering
	\includegraphics[width=\linewidth]{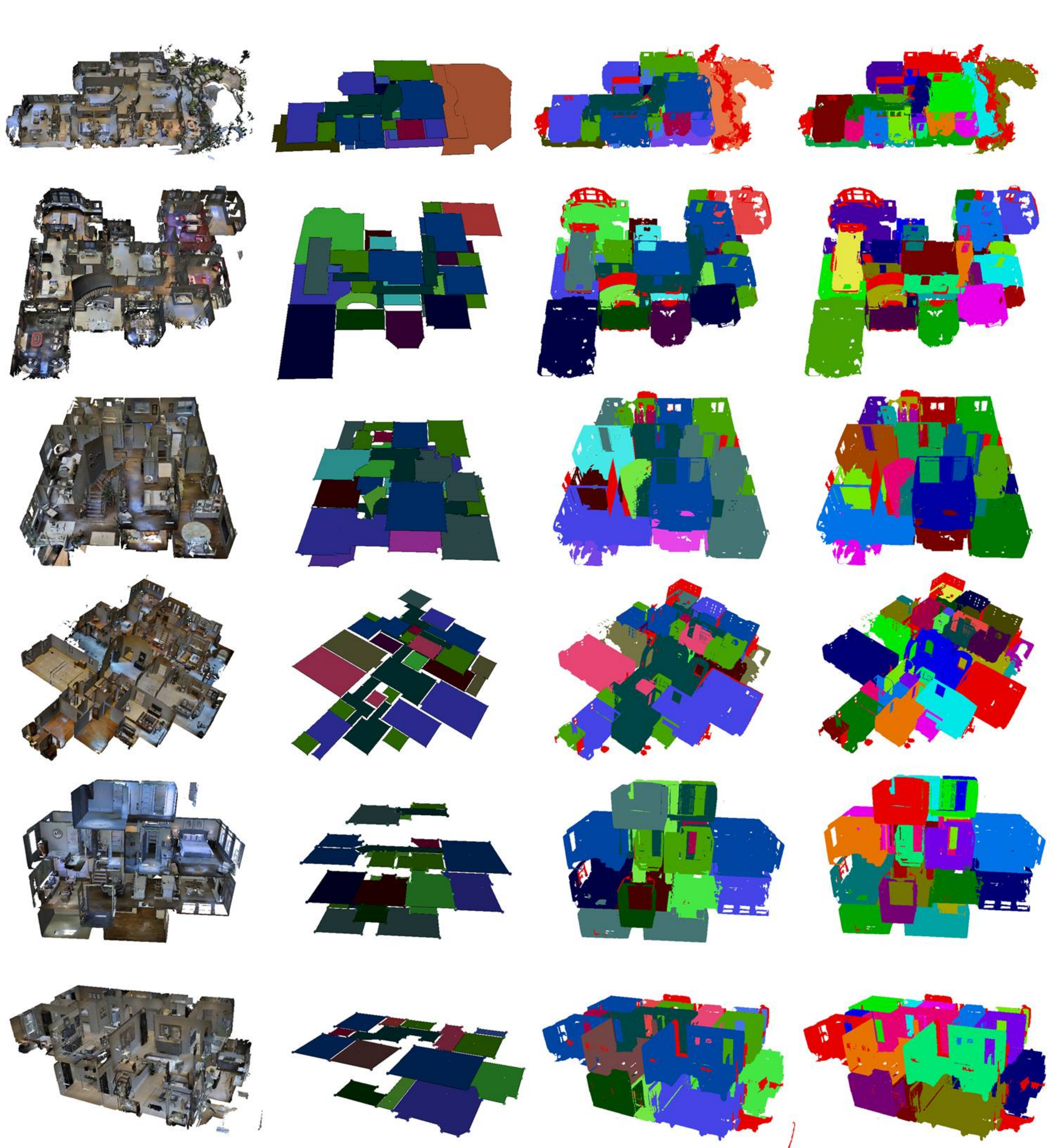} \\
	\hfill a) Textured mesh \hfill  b) Floorplan \hfill c) Region semantic labels \hfill  d) Region segmentation
	\caption{{\bf Region-type classification.} The dataset includes manually-specified boundary and category annotations for all regions (rooms) of all buildings.  This figure shows for several examples (from left to right): the textured mesh, the floorplan colored by region category, the mesh surface colored by region category, and the mesh surface colored by region instance.}
	\label{fig:floorplans1}
\end{figure*}

\begin{figure*}[p]
	\centering
	\includegraphics[width=\linewidth]{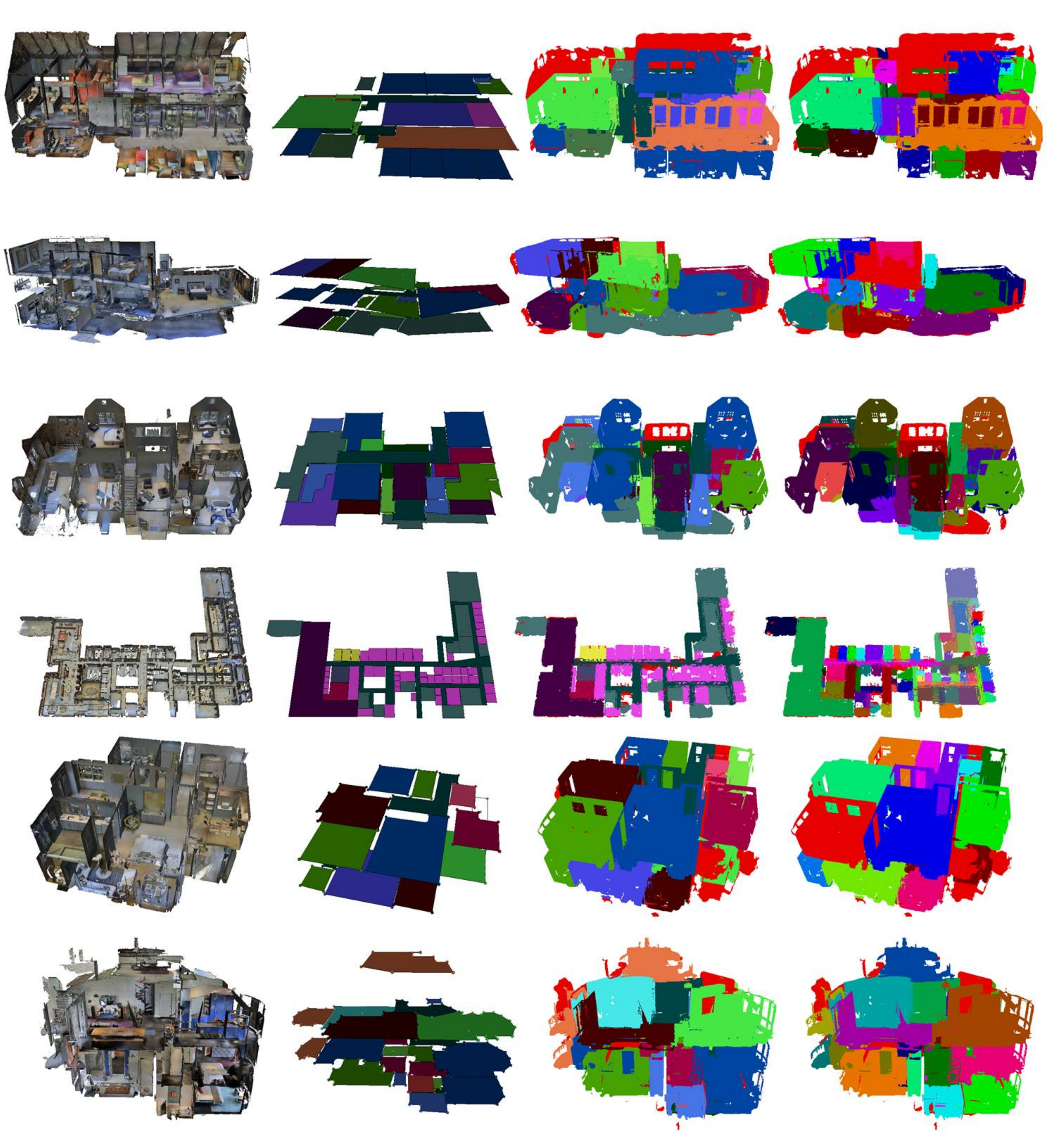} \\
	\hfill a) Textured mesh \hfill  b) Floorplan \hfill c) Region semantic labels \hfill  d) Region segmentation
	\caption{{\bf Region-type classification.} More examples like Fig. \ref{fig:floorplans1}.}
	\label{fig:floorplans2}
\end{figure*}

The fourth task is to predict the category of the region (room) containing a given panorama. 
Fig.~\ref{fig:floorplans1} and \ref{fig:floorplans2} show several examples of the region-type category annotations provided with \OURS.
For each building, a person manually outlined the floorplan boundary of each region (shown in column b) and provided a semantic label for its category (bathroom, bedroom, closet, dining room, entryway, familyroom, garage, hallway, library, laundryroom, kitchen, livingroom, meetingroom, lounge, office, porch, recroom, stairs, toilet, utilityroom, gym, outdoor, otherroom, bar, classroom, diningbooth, spa, or junk). 
Then, the imprecisely drawn region boundaries are snapped to the mesh surfaces and extruded to provide a full semantic segmentation of the original surface mesh by region category (c) and instance (d).  These category labels provide the ground truth for the region-type categorization task described in Section 4.4 of the main paper.

\subsection{Semantic Voxel Labeling}

\begin{figure*}
	\centering
	\begin{subfigure}[t]{0.49\linewidth}
		\centering
		Textured mesh\\
		\includegraphics[trim={4cm 0 17cm 0},clip,width=\linewidth]{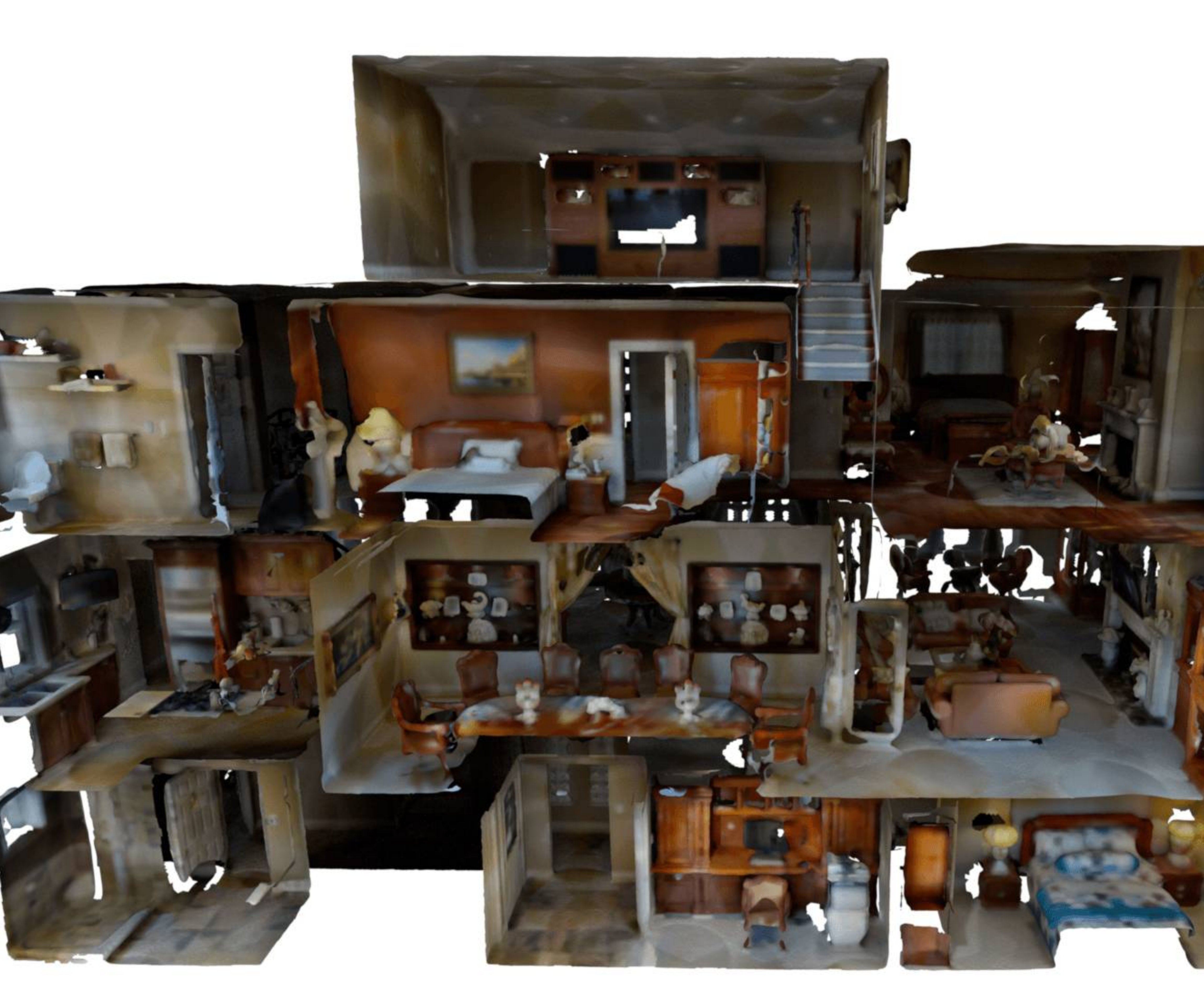}
	\end{subfigure}
	\begin{subfigure}[t]{0.49\linewidth}
		\centering
		Object instance labels\\
		\includegraphics[trim={4cm 0 17cm 0},clip,width=\linewidth]{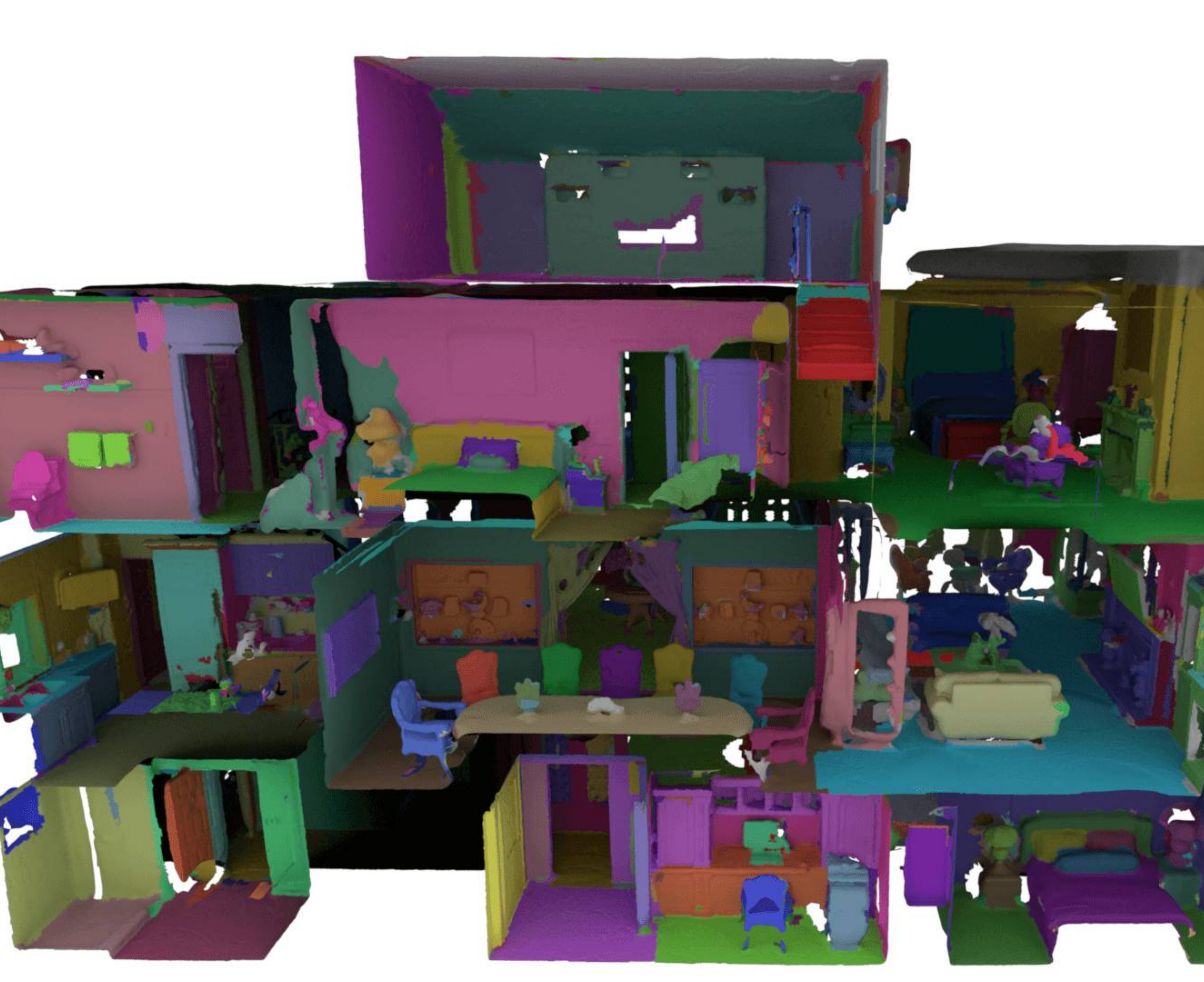}
	\end{subfigure}
	\vskip\baselineskip
	\vskip\baselineskip
	\begin{subfigure}[t]{0.49\linewidth}
		\centering
		Raw object category labels\\
		\includegraphics[trim={4cm 0 17cm 0},clip,width=\linewidth]{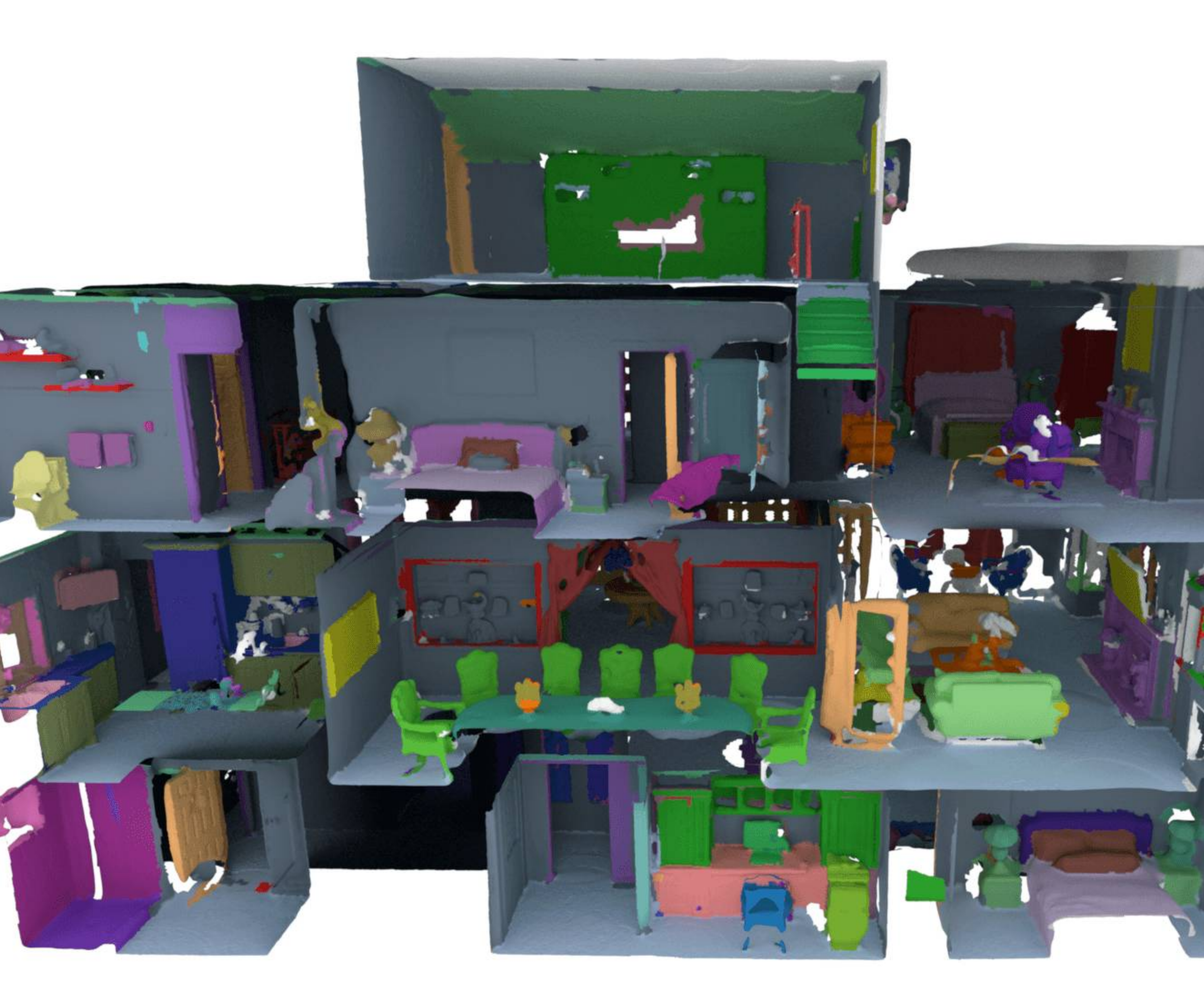}
	\end{subfigure}
	\begin{subfigure}[t]{0.49\linewidth}
		\centering
		Canonical 40 category object labels\\
		\includegraphics[trim={4cm 0 17cm 0},clip,width=\linewidth]{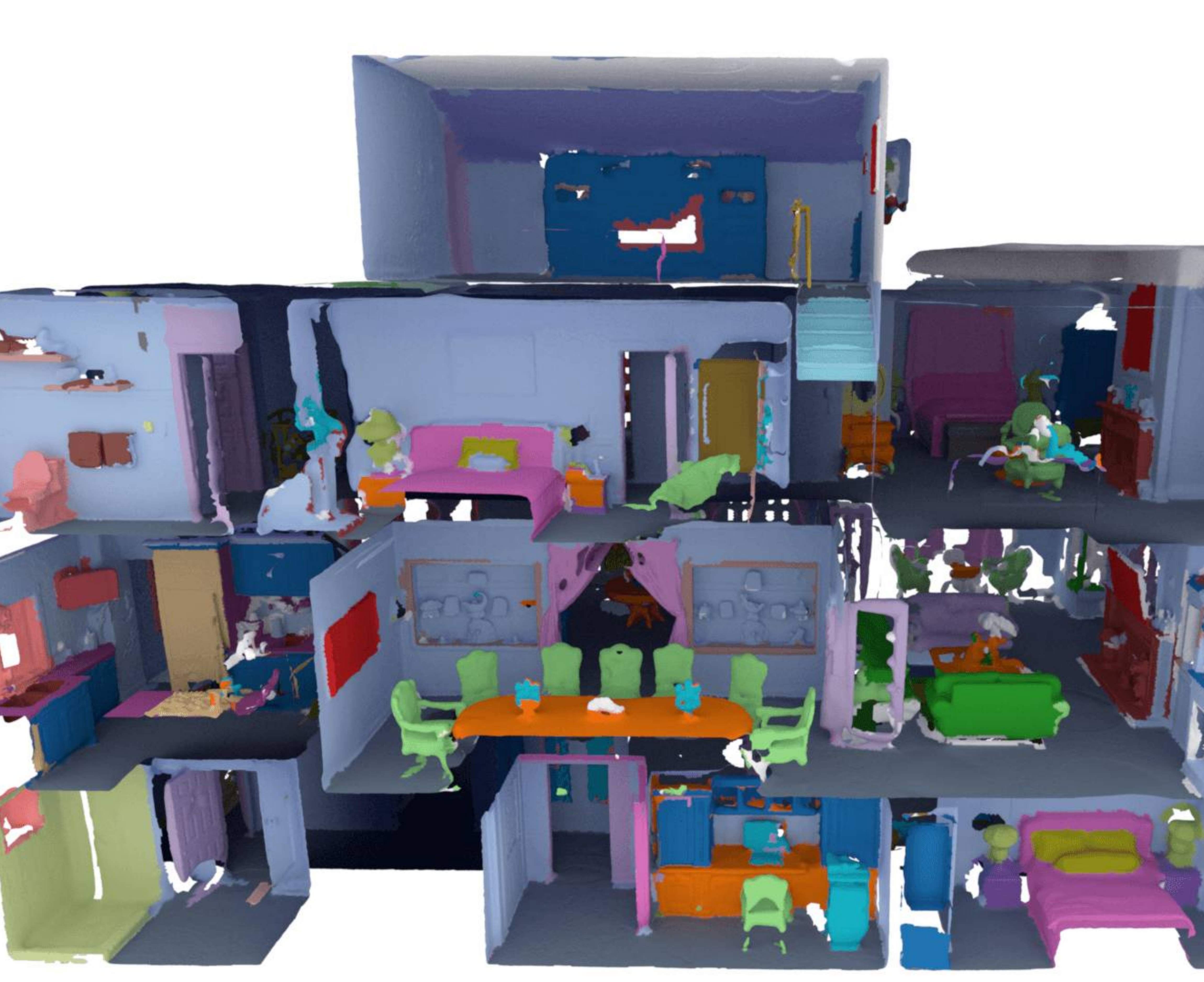}
	\end{subfigure}
	\caption{{\bf Semantic Voxel Label Segmentations.} The dataset includes manually ``painted'' object instance and category labels.  From top left: textured 3D mesh, object instances, object category labels, and finally canonicalized 40 category labels.  Note that raw labels for different types of chairs such as ``dining chair'', and ``office chair'' are mapped to a single canonical ``chair'' category shown in light green.}
	\label{fig:semantic_annotations}
\end{figure*}

The final task is to predict per-voxel labels for a given scene voxelization.  
Fig.~\ref{fig:semantic_annotations} shows an example house with instance-level semantic voxel annotations indicated by the colors.  
First, we partition the textured mesh of every house into region meshes as described in the previous section.  
Then, we obtain annotations for the object instances in each region using the crowdsourced semantic ``paint and name'' interface of Dai \etal \cite{dai2017scannet}; see Fig.~\ref{fig:semantic_annotations}, top right. 
We then apply spell checking and flattening of synonyms to obtain raw object category labels (Fig.~\ref{fig:semantic_annotations}, bottom left).  
Using occurrence frequency sorting and collapsing categories that are refinements of other categories (e.g., ``dining chair'' and ``chair''), we further reduce these labels to a canonical set of 40 categories (Fig.~\ref{fig:semantic_annotations}, bottom right).  
The canonical category labels constitute the ground truth for the semantic voxel labeling task described in Section 4.5 of the main paper.


%% file: main.bbl
\begin{thebibliography}{10}\itemsep=-1pt

\bibitem{armeni2017joint}
I.~Armeni, S.~Sax, A.~R. Zamir, and S.~Savarese.
\newblock Joint {2D-3D}-semantic data for indoor scene understanding.
\newblock {\em arXiv preprint arXiv:1702.01105}, 2017.

\bibitem{armeni20163d}
I.~Armeni, O.~Sener, A.~R. Zamir, H.~Jiang, I.~Brilakis, M.~Fischer, and
  S.~Savarese.
\newblock {3D} semantic parsing of large-scale indoor spaces.
\newblock In {\em Proceedings of the IEEE Conference on Computer Vision and
  Pattern Recognition}, pages 1534--1543, 2016.

\bibitem{bansal2016marr}
A.~Bansal, B.~Russell, and A.~Gupta.
\newblock Marr revisited: {2D-3D} alignment via surface normal prediction.
\newblock In {\em Proceedings of the IEEE Conference on Computer Vision and
  Pattern Recognition}, pages 5965--5974, 2016.

\bibitem{bell14intrinsic}
S.~Bell, K.~Bala, and N.~Snavely.
\newblock Intrinsic images in the wild.
\newblock {\em ACM Trans. on Graphics (SIGGRAPH)}, 33(4), 2014.

\bibitem{choi2016large}
S.~Choi, Q.-Y. Zhou, S.~Miller, and V.~Koltun.
\newblock A large dataset of object scans.
\newblock {\em arXiv preprint arXiv:1602.02481}, 2016.

\bibitem{chuang2011interactive}
M.~Chuang and M.~Kazhdan.
\newblock Interactive and anisotropic geometry processing using the screened
  poisson equation.
\newblock {\em ACM Transactions on Graphics (TOG)}, 30(4):57, 2011.

\bibitem{dai2017scannet}
A.~Dai, A.~X. Chang, M.~Savva, M.~Halber, T.~Funkhouser, and M.~Nie{\ss}ner.
\newblock Scannet: Richly-annotated 3d reconstructions of indoor scenes.
\newblock {\em http://arxiv.org/abs/1702.04405}, 2017.

\bibitem{dai2017bundle}
A.~Dai, M.~Nie{\ss}ner, M.~Zoll{\"o}fer, S.~Izadi, and C.~Theobalt.
\newblock Bundlefusion: Real-time globally consistent 3d reconstruction using
  on-the-fly surface re-integration.
\newblock {\em ACM Transactions on Graphics 2017 (TOG)}, 2017.

\bibitem{eigen2015predicting}
D.~Eigen and R.~Fergus.
\newblock Predicting depth, surface normals and semantic labels with a common
  multi-scale convolutional architecture.
\newblock In {\em Proceedings of the IEEE International Conference on Computer
  Vision}, pages 2650--2658, 2015.

\bibitem{firman2016rgbd}
M.~Firman.
\newblock {RGBD} datasets: Past, present and future.
\newblock In {\em CVPR Workshop on Large Scale 3D Data: Acquisition, Modelling
  and Analysis}, 2016.

\bibitem{fouhey2013data}
D.~F. Fouhey, A.~Gupta, and M.~Hebert.
\newblock Data-driven {3D} primitives for single image understanding.
\newblock In {\em ICCV}, 2013.

\bibitem{fouhey2014unfolding}
D.~F. Fouhey, A.~Gupta, and M.~Hebert.
\newblock Unfolding an indoor origami world.
\newblock In {\em European Conference on Computer Vision}, pages 687--702.
  Springer, 2014.

\bibitem{gupta2013perceptual}
S.~Gupta, P.~Arbelaez, and J.~Malik.
\newblock Perceptual organization and recognition of indoor scenes from {RGB-D}
  images.
\newblock In {\em Proceedings of the IEEE Conference on Computer Vision and
  Pattern Recognition}, pages 564--571, 2013.

\bibitem{gupta2014learning}
S.~Gupta, R.~Girshick, P.~Arbel{\'a}ez, and J.~Malik.
\newblock Learning rich features from {RGB-D} images for object detection and
  segmentation: Supplementary material, 2014.

\bibitem{StructuredGlobalRegistration}
M.~Halber and T.~Funkhouser.
\newblock Structured global registration of rgb-d scans in indoor environments.
\newblock 2017.

\bibitem{han2015matchnet}
X.~Han, T.~Leung, Y.~Jia, R.~Sukthankar, and A.~C. Berg.
\newblock Matchnet: Unifying feature and metric learning for patch-based
  matching.
\newblock In {\em Proceedings of the IEEE Conference on Computer Vision and
  Pattern Recognition}, pages 3279--3286, 2015.

\bibitem{handa2015scenenet}
A.~Handa, V.~Patraucean, V.~Badrinarayanan, S.~Stent, and R.~Cipolla.
\newblock Scene{N}et: Understanding real world indoor scenes with synthetic
  data.
\newblock {\em arXiv preprint arXiv:1511.07041}, 2015.

\bibitem{he2016deep}
K.~He, X.~Zhang, S.~Ren, and J.~Sun.
\newblock Deep residual learning for image recognition.
\newblock In {\em Proceedings of the IEEE Conference on Computer Vision and
  Pattern Recognition}, pages 770--778, 2016.

\bibitem{hoffer2016deep}
E.~Hoffer, I.~Hubara, and N.~Ailon.
\newblock Deep unsupervised learning through spatial contrasting.
\newblock {\em arXiv preprint arXiv:1610.00243}, 2016.

\bibitem{hua2016scenenn}
B.-S. Hua, Q.-H. Pham, D.~T. Nguyen, M.-K. Tran, L.-F. Yu, and S.-K. Yeung.
\newblock {SceneNN}: A scene meshes dataset with annotations.
\newblock In {\em International Conference on 3D Vision (3DV)}, volume~1, 2016.

\bibitem{kazhdan2006poisson}
M.~Kazhdan, M.~Bolitho, and H.~Hoppe.
\newblock Poisson surface reconstruction.
\newblock In {\em Proceedings of the fourth Eurographics symposium on Geometry
  processing}, volume~7, 2006.

\bibitem{knapitsch2017tanks}
A.~Knapitsch, J.~Park, Q.-Y. Zhou, and V.~Koltun.
\newblock Tanks and temples: Benchmarking large-scale scene reconstruction.
\newblock {\em ACM Transactions on Graphics}, 36(4), 2017.

\bibitem{li2015depth}
B.~Li, C.~Shen, Y.~Dai, A.~van~den Hengel, and M.~He.
\newblock Depth and surface normal estimation from monocular images using
  regression on deep features and hierarchical {CRF}s.
\newblock In {\em Proceedings of the IEEE Conference on Computer Vision and
  Pattern Recognition}, pages 1119--1127, 2015.

\bibitem{lin2013holistic}
D.~Lin, S.~Fidler, and R.~Urtasun.
\newblock Holistic scene understanding for {3D} object detection with rgbd
  cameras.
\newblock In {\em Proceedings of the IEEE International Conference on Computer
  Vision}, pages 1417--1424, 2013.

\bibitem{niessner2013hashing}
M.~Nie{\ss}ner, M.~Zollh\"ofer, S.~Izadi, and M.~Stamminger.
\newblock Real-time {3D} reconstruction at scale using voxel hashing.
\newblock {\em ACM Transactions on Graphics (TOG)}, 2013.

\bibitem{rematas2016deep}
K.~Rematas, T.~Ritschel, M.~Fritz, E.~Gavves, and T.~Tuytelaars.
\newblock Deep reflectance maps.
\newblock In {\em Proceedings of the IEEE Conference on Computer Vision and
  Pattern Recognition}, pages 4508--4516, 2016.

\bibitem{ren2012rgb}
X.~Ren, L.~Bo, and D.~Fox.
\newblock {RGB-(D)} scene labeling: Features and algorithms.
\newblock In {\em Computer Vision and Pattern Recognition (CVPR), 2012 IEEE
  Conference on}, pages 2759--2766. IEEE, 2012.

\bibitem{savva2016pigraphs}
M.~Savva, A.~X. Chang, P.~Hanrahan, M.~Fisher, and M.~Nie{\ss}ner.
\newblock {PiGraphs: Learning Interaction Snapshots from Observations}.
\newblock {\em ACM Transactions on Graphics (TOG)}, 35(4), 2016.

\bibitem{schmidt2017self}
T.~Schmidt, R.~Newcombe, and D.~Fox.
\newblock Self-supervised visual descriptor learning for dense correspondence.
\newblock {\em IEEE Robotics and Automation Letters}, 2(2):420--427, 2017.

\bibitem{shotton2013scene}
J.~Shotton, B.~Glocker, C.~Zach, S.~Izadi, A.~Criminisi, and A.~Fitzgibbon.
\newblock Scene coordinate regression forests for camera relocalization in
  {RGB-D} images.
\newblock In {\em Proceedings of the IEEE Conference on Computer Vision and
  Pattern Recognition}, pages 2930--2937, 2013.

\bibitem{shrivastava2013building}
A.~Shrivastava and A.~Gupta.
\newblock Building part-based object detectors via {3D} geometry.
\newblock In {\em Proceedings of the IEEE International Conference on Computer
  Vision}, pages 1745--1752, 2013.

\bibitem{silberman2011indoor}
N.~Silberman and R.~Fergus.
\newblock Indoor scene segmentation using a structured light sensor.
\newblock In {\em Computer Vision Workshops (ICCV Workshops), 2011 IEEE
  International Conference on}, 2011.

\bibitem{silberman2012indoor}
N.~Silberman, D.~Hoiem, P.~Kohli, and R.~Fergus.
\newblock Indoor segmentation and support inference from {RGBD} images.
\newblock In {\em European Conference on Computer Vision}, 2012.

\bibitem{simo2015discriminative}
E.~Simo-Serra, E.~Trulls, L.~Ferraz, I.~Kokkinos, P.~Fua, and F.~Moreno-Noguer.
\newblock Discriminative learning of deep convolutional feature point
  descriptors.
\newblock In {\em Proceedings of the IEEE International Conference on Computer
  Vision}, pages 118--126, 2015.

\bibitem{song2015sun}
S.~Song, S.~P. Lichtenberg, and J.~Xiao.
\newblock {SUN RGB-D}: A {RGB-D} scene understanding benchmark suite.
\newblock In {\em Proceedings of the IEEE conference on computer vision and
  pattern recognition}, pages 567--576, 2015.

\bibitem{song2014sliding}
S.~Song and J.~Xiao.
\newblock Sliding shapes for {3D} object detection in depth images.
\newblock In {\em European conference on computer vision}, pages 634--651.
  Springer, 2014.

\bibitem{song2015deep}
S.~Song and J.~Xiao.
\newblock Deep sliding shapes for amodal {3D} object detection in {RGB-D}
  images.
\newblock 2016.

\bibitem{song2016semantic}
S.~Song, F.~Yu, A.~Zeng, A.~X. Chang, M.~Savva, and T.~Funkhouser.
\newblock Semantic scene completion from a single depth image.
\newblock {\em arXiv preprint arXiv:1611.08974}, 2016.

\bibitem{valentin2016learning}
J.~Valentin, A.~Dai, M.~Nie{\ss}ner, P.~Kohli, P.~Torr, S.~Izadi, and
  C.~Keskin.
\newblock Learning to navigate the energy landscape.
\newblock {\em arXiv preprint arXiv:1603.05772}, 2016.

\bibitem{valentin2015semanticpaint}
J.~Valentin, V.~Vineet, M.-M. Cheng, D.~Kim, J.~Shotton, P.~Kohli,
  M.~Nie{\ss}ner, A.~Criminisi, S.~Izadi, and P.~Torr.
\newblock {SemanticPaint}: Interactive {3D} labeling and learning at your
  fingertips.
\newblock {\em ACM Transactions on Graphics (TOG)}, 34(5):154, 2015.

\bibitem{wang2015designing}
X.~Wang, D.~Fouhey, and A.~Gupta.
\newblock Designing deep networks for surface normal estimation.
\newblock In {\em Proceedings of the IEEE Conference on Computer Vision and
  Pattern Recognition}, pages 539--547, 2015.

\bibitem{xiao2012recognizing}
J.~Xiao, K.~A. Ehinger, A.~Oliva, and A.~Torralba.
\newblock Recognizing scene viewpoint using panoramic place representation.
\newblock In {\em Computer Vision and Pattern Recognition (CVPR), 2012 IEEE
  Conference on}, pages 2695--2702. IEEE, 2012.

\bibitem{xiao2010sun}
J.~Xiao, J.~Hays, K.~A. Ehinger, A.~Oliva, and A.~Torralba.
\newblock Sun database: Large-scale scene recognition from abbey to zoo.
\newblock In {\em Computer vision and pattern recognition (CVPR), 2010 IEEE
  conference on}, pages 3485--3492. IEEE, 2010.

\bibitem{xiao2013sun3d}
J.~Xiao, A.~Owens, and A.~Torralba.
\newblock {SUN3D}: A database of big spaces reconstructed using {SFM} and
  object labels.
\newblock In {\em Proceedings of the IEEE International Conference on Computer
  Vision}, pages 1625--1632, 2013.

\bibitem{yi2016lift}
K.~M. Yi, E.~Trulls, V.~Lepetit, and P.~Fua.
\newblock {LIFT}: Learned invariant feature transform.
\newblock In {\em European Conference on Computer Vision}, pages 467--483.
  Springer, 2016.

\bibitem{zeng20163dmatch}
A.~Zeng, S.~Song, M.~Niessner, M.~Fisher, J.~Xiao, and T.~Funkhouser.
\newblock {3DMatch}: Learning local geometric descriptors from {RGB-D}
  reconstructions.
\newblock {\em Proceedings of the IEEE Conference on Computer Vision and
  Pattern Recognition}, 2017.

\bibitem{zhang2013estimating}
J.~Zhang, C.~Kan, A.~G. Schwing, and R.~Urtasun.
\newblock Estimating the {3D} layout of indoor scenes and its clutter from
  depth sensors.
\newblock In {\em Proceedings of the IEEE International Conference on Computer
  Vision}, pages 1273--1280, 2013.

\bibitem{zhang2016physically}
Y.~Zhang, S.~Song, E.~Yumer, M.~Savva, J.-Y. Lee, H.~Jin, and T.~Funkhouser.
\newblock Physically-based rendering for indoor scene understanding using
  convolutional neural networks.
\newblock {\em arXiv preprint arXiv:1612.07429}, 2016.

\bibitem{zhou2014learning}
B.~Zhou, A.~Lapedriza, J.~Xiao, A.~Torralba, and A.~Oliva.
\newblock Learning deep features for scene recognition using places database.
\newblock In {\em Advances in neural information processing systems}, pages
  487--495, 2014.

\end{thebibliography}
